\documentclass{article}

\usepackage[preprint]{neurips_2025}

\usepackage[utf8]{inputenc} 
\usepackage[T1]{fontenc}    
\usepackage{hyperref}       
\usepackage{url}            
\usepackage{booktabs}       
\usepackage{amsfonts}       
\usepackage{nicefrac}       
\usepackage{microtype}      
\usepackage{xcolor}  
\usepackage{soul}
\usepackage{tcolorbox}

\usepackage{graphicx}   
\usepackage{subcaption} 
\usepackage{array}      
\usepackage{colortbl}   

\usepackage[table]{xcolor}
\definecolor{pelicanmain}{HTML}{FBE6B2}
\definecolor{pelicanlight}{HTML}{FFF3D6}
\definecolor{myred}{RGB}{220, 20, 60}

\usepackage{multirow}
\usepackage{multicol}
\usepackage{graphicx}
\usepackage{wrapfig}
\usepackage{algorithm}
\usepackage{algpseudocode}
\usepackage{changepage}

\usepackage{amssymb, amsmath}
\usepackage{dsfont}
\usepackage{cleveref}



\title{Bridging VLMs and Embodied Intelligence with \\ Deliberate Practice Policy Optimization}

\author{%
Yi Zhang$^{1,*}$,
Che Liu$^{2,*}$,
Xiancong Ren$^{1,*}$,
Hanchu Ni$^{3}$,
Yingji Zhang$^{4}$,
Shuai Zhang$^{5}$,
Zeyuan Ding$^{1}$,\\
\textbf{Jiayu Hu}$^{1}$,
\textbf{Haozhe Shan}$^{6}$,
\textbf{Junbo Qi}$^{7}$,
\textbf{Yan Bai}$^{8}$,
\textbf{Dengjie Li}$^{1}$,
\textbf{Jiachen Luo}$^{9}$,
\textbf{Yidong Wang}$^{3}$,\\
\textbf{Yong Dai}$^{1,\ddagger}$,
\textbf{Zenglin Xu}$^{6}$,
\textbf{Bin Shen}$^{10}$,
\textbf{Qifan Wang}$^{11}$,
\textbf{Jian Tang}$^{1,\triangledown}$,
\textbf{Xiaozhu Ju}$^{1,\triangledown}$\\[2mm]
\small
$^{1}$X-Humanoid, $^{2}$Imperial College London, $^{3}$Peking University, $^{4}$University of Manchester\\
$^{5}$Westlake University, $^{6}$Fudan University, $^{7}$Waseda University, $^{8}$Nvidia\\
$^{9}$Queen Mary University of London, $^{10}$Celonis AI, $^{11}$Meta AI\\
$^{*}$Joint first author,
$^{\ddagger}$Project lead,
$^{\triangledown}$Joint last author \\
    \textcolor{myred}{\textbf{Project Website:}} \href{https://pelican-vl.github.io/}{[W]}
    \quad
    \textcolor{myred}{\textbf{Models:}} \href{https://huggingface.co/X-Humanoid}{[M]}
}

\begin{document}

\maketitle

\begin{abstract}

\begin{tcolorbox}[
    colback=gray!15,      
    colframe=white!0,    
    boxrule=0pt,         
    arc=2mm,             
    left=4mm,            
    right=4mm,           
    width=\textwidth,    
    enlarge left by=0mm, 
    enlarge right by=0mm
]
Developing a universal and versatile embodied intelligence system presents two primary challenges: the critical embodied data bottleneck, where real-world data is scarce and expensive, and the algorithmic inefficiency of existing methods, which are resource-prohibitive. To address these limitations, we introduce Deliberate Practice Policy Optimization (DPPO), a metacognitive ``Metaloop'' training framework that dynamically alternates between supervised fine-tuning (competence expansion) and reinforcement learning (skill refinement). This enables automatic weakness identification and targeted resource allocation, specifically designed to maximize learning efficiency from sparse, finite data. Theoretically, DPPO can be formalised as a unified preference-learning framework. Empirically, training a vision-language embodied model with DPPO, referred to as Pelican-VL 1.0, yields a 20.3\% performance improvement over the base model and surpasses open-source models at the 100B-parameter scale by 10.6\%. We are open-sourcing both the models and code, providing the first systematic framework that alleviates the data and resource bottleneck and enables the community to build versatile embodied agents efficiently.
\end{tcolorbox}

\end{abstract}

\section{Introduction}

Artificial General Intelligence (AGI) requires embodied intelligence systems capable of integrating a comprehensive spectrum of abilities, including perception, reasoning, cross-modal understanding, and real-world interaction~\cite{khan2025foundation,xiang2025parallels,liu2025nexus}. In pursuit of such generalist embodied intelligence systems, recent research leveraging increasingly large-scale models has generally bifurcated into two complementary strategies~\cite{brohan2023rt,team2025gemini,driess2023palm,ahn2022can}: \textit{(1) heterogeneous data scaling:} models such as Gemini Robotics~\cite{team2025gemini}, GR00T N1, $\pi_{0.5}$~\cite{intelligence2504pi0}, and GR-3~\cite{abdolmaleki2025gemini,bjorck2025gr00t,cheang2025gr} construct massive ``data pyramids'' by fusing web, simulation, and real-world trajectories~\cite{khan2025foundation}; and \textit{(2) architectural refinement for control:} approaches like Helix~\cite{Helix_Figure_2025} and Wall-OSS~\cite{zhai2025igniting} improve continuous control for high-DoF robots. Despite their rapid progress, these strategies do not fundamentally resolve the embodiment bottleneck. Both remain extensions of the offline imitation learning paradigm, scaling passive data volume (Strategy 1) or control fidelity (Strategy 2) without improving true data efficiency or enabling metacognitive self-improvement~\cite{flavell1979metacognition,brown1987metacognition}. Critically, neither provides a mechanism for continual adaptation: models are trained in a one-shot, offline manner and cannot autonomously identify weaknesses or refine skills based on sparse, targeted feedback during real-world deployment.

In contrast, humans leverage \textit{metacognition}, the ability to monitor, evaluate, and regulate their own learning, to learn efficiently by actively identifying weaknesses, selectively allocating effort, and refining strategies through feedback. Inspired by this principle, we introduce \textbf{computational metacognition} as a paradigm that shifts learning from passive data accumulation to active competence expansion through self-diagnosis and self-refinement.
\begin{figure*}
    \centering
    \includegraphics[width=1\linewidth]{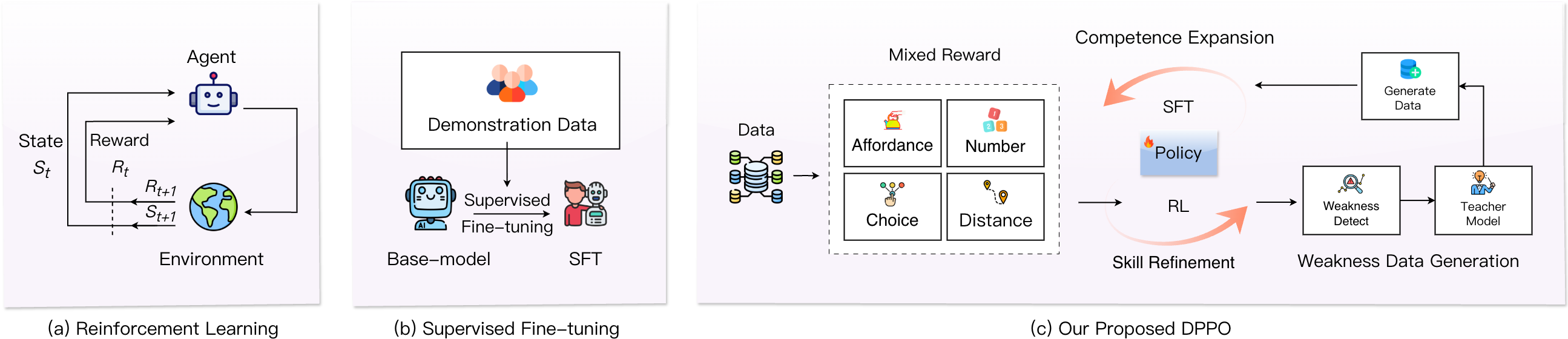}
        \caption{Overview of DPPO. The framework implements an iterative RL–SFT metaloop that leverages rollout logging and difficulty-aware sampling for dynamic data curation. This adaptive process alternates between \textbf{revealing weaknesses} in the RL phase and \textbf{refining them} in the SFT phase, forming a continual self-diagnosis and self-refinement cycle.} \label{fig:overview}
\end{figure*}

We operationalize this paradigm through \textbf{Deliberate Practice Policy Optimization (DPPO)}, a metacognitive \textit{metaloop} that dynamically alternates between two synergistic stages of exploration and consolidation. In this framework, the RL phase uncovers model weaknesses through exploratory interaction, while the SFT phase refines and strengthens these weaknesses via guided supervision. Together, these stages constitute a continual process of \textit{self-diagnosis and self-refinement}, transforming sparse, failure-driven experience into systematic capability growth.

Operationally, DPPO proceeds as follows: (1) the RL-trained policy performs rollouts that are diagnostically monitored to automatically identify hard cases (e.g., tasks showing consistent failure patterns). (2) These weaknesses are provided to a teacher model (e.g., InternVL 3.5~\cite{wang2025internvl3_5}), which generates high-quality reference solutions. (3) The SFT phase then distils these solutions through supervised learning to expand competence and strengthen refined skills. This targeted feedback loop alleviates data imbalance and enables efficient learning from limited, high-value experience. Theoretically, the entire process is grounded in a unified preference-learning formulation, revealing the synergy between weakness revelation (RL) and refinement (SFT).

We validate DPPO by training a vision–language model (Qwen2.5-VL~\cite{Yang2024Qwen25TR}) on a data-efficient corpus. our  72b model achieves a 20.3\% gain over its base model and surpasses open-source models at the 100B-parameter scale by 10.6\%. The main contributions are as follows:
\begin{itemize}
    \item \textbf{DPPO Framework.} DPPO unifies RL for \textit{weakness revelation} and SFT for \textit{weakness refinement} within an adaptive \textit{self-diagnosis and self-refinement} metaloop. It continuously identifies weaknesses and allocates learning resources efficiently, and can be theoretically framed as unified preference learning.
\item \textbf{Training Effectiveness.} DPPO serves as an intelligent data engine: RL reveals challenging cases, converting sparse data into high-value signals, while SFT consolidates and generalizes them.  Pelican-VL 1.0 72B achieve 20.3\% higher performance.
\item \textbf{Open Source.} We release Pelican-VL 1.0 (7B–72B) and the complete DPPO pipeline to advance scalable, capital-efficient, self-improving embodied intelligence.

\end{itemize}

\section{Related Work}
\label{sec:relatedwork}

\paragraph{Post-training algorithm.} 
Post-training algorithms are essential for aligning foundational models. The baseline approach, SFT~\cite{tie2025survey,gui2024survey}, is a stable mechanism for static knowledge infusion from expert data. While effective for domain adaptation, it is inherently limited by its pre-collected dataset and cannot discover novel policies. This limitation led to the integration of RL~\cite{wu2025reinforced,zhai2024rl4vlm} algorithms. The field evolved from complex, online reward-modeling methods like PPO~\cite{schulman2017proximal} to more stable, offline alignment techniques~\cite{cheng2024adversarial}. DPO~\cite{rafailov2023direct} marked a significant algorithmic shift by reframing alignment as a direct preference-learning problem. This was further refined by algorithms like GRPO~\cite{shao2024deepseekmath}, which generalizes DPO to leverage richer, ranked preference signals instead of just binary pairs.

Beyond these baselines, recent hybrid or unified post-training frameworks explore tighter couplings between supervised signals and RL or propose stabilized policy-optimization objectives, e.g., POLAR~\cite{Dou2025PreTrainedPD}, Hybrid Post-Training (HPT)~\cite{Lv2025TowardsAU}, CHORD~\cite{Zhang2025OnPolicyRM}, implicit reward–based alignment~\cite{Wang2025ImplicitRA}, Dynamic Fine-Tuning (DFT)~\cite{Wu2025OnTG}, Unified Fine-Tuning (UFT)~\cite{Liu2025UFTUS,qian2025uniapl}, and Soft Analytic Policy Optimization (SAPO)~\cite{Xing2024StabilizingRL}. Our work differs by operationalizing this coupling as a metaloop that alternates discovery (RL) and consolidation (SFT), turning weakness detection into targeted supervision rather than a one-off pipeline.

\paragraph{Data selection algorithm.} 
Data selection has traditionally been formulated as a coreset selection problem, where the goal is to identify a small but representative subset of training samples that achieves similar performance to full-dataset training \cite{Vegter1997InHO,Toneva2018AnES,Mindermann2022PrioritizedTO,Sorscher2022BeyondNS,Xia2023ModerateCA}. Methods include selecting based on pre-defined features \cite{Gururangan2020DontSP,Chen2023SkillitAD}, n-gram features \cite{Xie2023DataSF}, or
gradient-based features \cite{Wang2019OptimizingDU,mirzasoleiman2020coresetsdataefficienttrainingmachine,Yu2020GradientSF,Killamsetty2021GRADMATCHGM,xia2024lessselectinginfluentialdata}. While effective, most of these methods assume in-domain off-line coreset selection, i.e., sample ranking within a fixed dataset, and do not model iterative skill acquisition across stages of training. Follow-ups in active/online selection \cite{Settles2009ActiveLL,Kirsch2019BatchBALDEA} and curriculum/hard-example mining \cite{Bengio2009CurriculumL,Milano2021AutomatedCL,Lin2017FocalLF} update the pool during optimization, yet still depend on proxy scores not explicitly tied to evolving capability gaps. Embodied settings add further challenges (video/trajectory length, multi-objective rewards, sparse successes), where dataset-agnostic scores can misalign with the skills that actually limit performance \cite{eppner2021acronym,gu2023maniskill2,dasari2019robonet,walke2023bridgedata,vuong2023open,bu2025agibot,qu2025embodiedonevision,shang2025roboscape}.

Unlike traditional methods, our approach does not rely on predefined heuristics or static gradient scoring or fixed data pool. DPPO performs dynamic, closed-loop data selection: during RL exploration, rollout trajectories are ranked by difficulty and only the hardest samples are injected into the subsequent SFT stage. This enables the model to continually identify its own failure modes and request supervision on the capability gaps that hinder learning.

\section{Method}
\label{sec:method}

In this section, we first formalize the embodied learning problem. We then present the DPPO framework, a metaloop engine that implements a self-evolving RL–SFT cycle to iteratively enhance embodied competencies. Finally, we establish the theoretical underpinnings of DPPO, demonstrating that SFT and RL can be unified under a coherent preference-learning framework.

\begin{algorithm}[t]
\caption{DPPO Metaloop}
\label{alg:dppo}
\begin{algorithmic}[1]
\Require Initial model parameters $\theta_0$, initial difficulty-aware buffer $\mathcal{B}$, 
saturation threshold $S_{\text{thresh}}$ (e.g., 0.7), number of iterations $K$.

\For{$k = 1$ to $K$}

    \Statex \textbf{Phase 1: RL for Weakness Detection (Sec.~\ref{sec:rl})}
    \Statex \textit{RL Data Construction}
    \For{each sample $c \in \mathcal{C}_k$}
    \State Generate trajectories $\mathcal{T}_k = \{\tau_1, \ldots, \tau_N\}$ via policy $\pi_{\theta_k}$.
    \State Compute SuccessRate score $S_R(c)$ by Eq.~\ref{eq:sr}.
    \State Add $(c, S_R(c))$ to buffer $\mathcal{B}$.
    \EndFor
    \State Build $\mathcal{D}^{k}_{RL}$ by re-balancing process in Section~\ref{sec:difficulty}

    \Statex \textit{Skill Refinement (on non-saturated tasks)}
    \For{mini-batch $\mathcal{D}_{\text{batch}} \sim \mathcal{D}^{k}_{RL}$}
        \State Update $\theta_k \leftarrow \theta_k - \eta \nabla_\theta \mathcal{L}_{\text{GRPO}}(\theta)$ (Eq.~\ref{eq:grpo}).
        \State Compute task stagnation score $S_S(task)$ Eq.~\ref{eq:task_stag}.
    \EndFor
    \State Build $\mathcal{D}_{weak}$ by rules in Section~\ref{sec:stop}
    \Statex \textbf{Phase 2: SFT for Weakness Refinement (Sec.~\ref{sec:sft})}
    \Statex \textit{SFT Data Construction (on saturated/hard tasks)}
    \State Build $\mathcal{D}_{\text{weak}}$ from $\mathcal{B}$ as mentioned before.
    \State Build $\mathcal{D}_{\text{rel}}$ by retrieving embodied-related samples.
    \State Build $\mathcal{D}_{\text{gen}}$ from general data for replay.
    \State $\mathcal{D}^{k}_{\text{SFT}} \leftarrow \mathcal{D}_{\text{weak}} \cup \mathcal{D}_{\text{rel}} \cup \mathcal{D}_{\text{gen}}$

    \Statex \textit{Competence Expansion}
    \For{each batch $\mathcal{D}_{\text{batch}} \in \mathcal{D}^{k}_{\text{SFT}}$}
        \State Update $\theta_k \leftarrow \theta_k - \eta \nabla_\theta \mathcal{L}_{\text{SFT}}(\theta)$ (Eq.~\ref{eq:sft}).
    \EndFor

    \Statex
    \State $\mathcal{B} \leftarrow \emptyset$ \Comment{Reset buffer to restart RL discovery cycle}

\EndFor
\State \Return Final model parameters $\theta_K$.
\end{algorithmic}
\end{algorithm}

\subsection{Problem Formulation: Embodied Scenarios and Tasks} \label{sec:problem_formulation}

This work formulates the problem as a composite multimodal task that integrates perception, scene understanding, task planning, physical manipulation, and human interaction, aligning with the core requirements of embodied intelligence systems~\cite{liu2024embodied}. Formally, given a multimodal input $(x_v, x_t)$ (i.e., visual frames and textual instructions) and a ground truth target $y$ (e.g., action sequences, reasoning traces, or function calls), our model $f_\theta$ is optimized by minimizing a task-appropriate loss function $\mathcal{L}(\theta)$:

\begin{equation}\label{eq:sft}
    \mathcal{L}(\theta) = \mathbb{E}_{(x_v, x_t, y)}[\ell(f_\theta(x_v, x_t), y)],
\end{equation}
where $\ell$ denotes the loss function, $v$ and $t$ denote the vision and text.

However, the simplified formulation actually reflects a complex real-world task such as causal-temporal inference or affordance reasoning. Consequently, single-stage training becomes hard to optimize both for SFT and RL: insufficient expert data prevents SFT from generalizing, while the large search space makes RL insufficiently constrained.

\subsection{Deliberate Practice Policy Optimization}\label{sec:dppo}
Inspired by the cognitive science principle of deliberate practice, we propose DPPO through a dynamically regulated metaloop. Rather than a monolithic optimization pipeline, the metaloop alternates between two synergistic phases: the RL phase that reveal flawed abilities through exploratory interaction, and the SFT phase that refines the weaknesses exposed by RL. Together, these alternating phases form a self-improving loop that enables the model to continually diagnose, target, and remediate embodied weaknesses by effectively harnessing and refining knowledge from large-scale, heterogeneous data.

Formally, let $\theta$ denote the model parameters and $k$ index the metaloop iterations. At each iteration, a \emph{phase selector} $\sigma_k \in \{\textsc{\textit{RL}},\,\textsc{\textit{SFT}}\}$ determines which objective is active, ensuring that the model optimizes only one phase-specific loss at a time:
\begin{equation}
\label{eq:phase-objective}
\mathcal{L}_{\sigma_k}(\theta) =
\begin{cases}
\mathcal{L}_{\textsc{RL}}(\theta; \mathcal{D}^{\,k}_{\text{RL}}) & \text{if } \sigma_k=\textsc{\textit{RL}}\\[4pt]
\mathcal{L}_{\textsc{SFT}}(\theta; \mathcal{D}^{k}_{\text{SFT}}) & \text{if } \sigma_k=\textsc{\textit{SFT}}.
\end{cases}
\end{equation}

Each full training cycle, $k$, consists of one RL phase followed by one SFT phase. The training process relies on distinct dynamic datasets for each phase, as defined in $\mathcal{L}_{\sigma_k}(\theta)$:

\begin{itemize}
    \item \textbf{For the RL phase ($\sigma_k=\textsc{\textit{RL}}$):} The objective $\mathcal{L}_{\textsc{RL}}$ uses the dataset $\mathcal{D}^{\,k}_{\text{RL}}$ which is dynamically constructed via difficulty-aware sampling from the original data pool, as detailed in Sec.~\ref{sec:rl}.

    \item \textbf{For the SFT phase ($\sigma_k=\textsc{\textit{SFT}}$):} The objective $\mathcal{L}_{\textsc{SFT}}$ uses $\mathcal{D}^{k}_{\text{SFT}}$, which is a union of three complementary \emph{dynamic} data sources, as detailed in Appendix ~\ref{sec:data_suppl}.
\end{itemize}

\subsubsection{Revealing Model Weaknesses via RL}\label{sec:rl}
In contrast to conventional RL that primarily focuses on maximizing cumulative rewards, this phase reframes RL as a process of exploratory diagnosis. Rather than solely pursuing reward maximization, the RL phase reveals latent weaknesses via exploratory interaction, allowing the model to uncover its limitations and identify areas for improvement. To support this process, we employ a GRPO framework equipped with multi-modal and multi-task reward functions, which guide the systematic discovery and characterization of such deficiencies in a data-driven manner. Formally, the policy gradient is expressed as:
\begin{equation}\label{eq:grpo}
  \nabla_\theta \mathcal{L}_{GRPO} = \mathbb{E}_{(x,y)\sim\pi_{ref}}[w(x,y) \nabla_\theta \log \pi_\theta(y|x)],
\end{equation}
where \(w(x, y)\) denotes the normalized reward weight derived from the rule-based scoring function that compares the current policy \(\pi_\theta\) against a reference policy \(\pi_{\text{ref}}\).

\vspace{4pt}
\noindent \textbf{Multi-modal and Multi-task Reward.}
To promote broad embodied competence, we construct a rule-based multi-task reward function covering six core objectives: affordance reasoning, counting and distance estimation, causal and temporal reasoning, task success evaluation, task planning, and task prediction. These objectives jointly guide the model to balance perception, reasoning, and planning abilities across multimodal embodied tasks, as detailed in Appendix~\ref{sec:method_suppl}. In addition, to stabilize both training and evaluation, we employ a format reward to constrain the model’s output. The format reward is computed by verifying whether the generated response strictly adheres to the required structure, specifically, that it contains a step-by-step reasoning trace and a correct final answer.

Each rollout $\tau$ receives a composite reward that combines a general format correctness term with a task-specific reward:

\begin{equation}\label{eq:format}
R(\tau) = \lambda_{\mathrm{f}}\, R_{\mathrm{f}}(\tau) + \lambda_{\mathrm{t}}\, R_{\mathrm{t}}(\tau)
\end{equation}
where $R_{\mathrm{f}}(\tau)$ measures the structural validity of the model’s output, and $R_{\mathrm{t}}(\tau)$ reflects the rule-based reward associated with the specific embodied objective, such as affordance reasoning or task planning. The coefficients $\lambda_{\mathrm{f}}$ and $\lambda_{\mathrm{t}}$ serve as weighting factors that balance the contribution of format correctness and task-level reward.

\vspace{4pt}
\noindent \textbf{Difficulty-aware sampling.}\label{sec:difficulty}
Before RL training, we analyse all samples to construct a difficulty-aware data buffer. This buffer consists of samples (e.g., specific task instances). For each sample $c$, we compute its SuccessRate score $S_R$ based on its aggregated rollout results:
\vspace{1pt}
\begin{equation}\label{eq:sr}
\mathrm{S_R}(c) = \frac{1}{\mathcal{T}} \sum_{i=1}^{\mathcal{T}} \mathbb{I}\!\left[\,R_i(c) = \mathrm{success}\,\right],
\end{equation}
where $R_i(c)$ is the outcome of the $i$-th rollout for sample $c$, and $\mathbb{I}(\cdot)$ is the indicator function, which equals 1 if the outcome is a success and 0 otherwise. Samples with a SuccessRate of 1 indicate tasks the model has already mastered and thus provide no additional learning signal, while a score of 0 indicates complete failure, which can hinder RL training since the model lacks the prerequisite knowledge. To foster more stable training and focus the model on meaningful signals, we pre-filter the dataset to construct the training dataset $\mathcal{D}_{\text{RL}}$ prior to initiating rl training phrase. This re-balancing process involves two steps:
\begin{itemize}
    \item We discard all samples with a 100\% success rate, as these tasks provide no learning signal.
    \item We cap the total count of complete failure samples ($S_R=0$), ensuring this count remains less than or equal to the total count of samples with partial success (i.e., those with $0 < S_R < 1$).
\end{itemize}

\noindent \textbf{RL training and stopping criterion.} \label{sec:stop}
During RL optimization, we track a stagnation score for each task to assess whether additional policy updates yield meaningful improvement. For a sample $c$, we define the SuccessRate  change \( \Delta(c) = \min\left(1, \frac{|S_R(c) - S_R^{\text{prev}}(c)|}{\varepsilon}\right) \), where  \( \varepsilon \in (0.05, 0.2) \) is chosen based on tasks to map \(\Delta(c)\) to \([0,1]\). For the sample $c$, the stagnation score $S_S$ is defined as:

\begin{equation}\label{eq:stag}
S_S(c) = 1 - 4S_R(c)(1 - S_R(c))\Delta(c),
\end{equation}\label{eq:sat}

\begin{itemize}
    \item If $S_R(c)$ is close to 0 or 1 (i.e., the sample is consistently failing or fully mastered), then $S_R(c)(1 - S_R(c)) \approx 0$, $S_S \approx 1$, indicating high stagnation.
    
    \item If the recent change is small (\(\Delta(c) \approx 0\)) despite \(S_R(c)\) not being near the extremes, then \(S_S \approx 1\) as well, reflecting stagnation due to lack of progress.
\end{itemize}

The task stagnation score $S_S(task)$ is obtained via aggregation over its associated samples:
\begin{equation}\label{eq:task_stag}
S_S(task) = \frac{1}{N_{\text{task}}} \sum_{i=1}^{N_{\text{task}}} S_S(c_i)
\end{equation}


Where $N_{task}$ denotes the number of samples in this task. The RL stage is automatically terminated when the stagnation score 
$S_S(task) \ge 0.7$ . 
At the end of each RL round, we further collect all samples with ${S_R=0}$
and include them in \( \mathcal{D}_{\text{weak}} \)  for the subsequent 
SFT updates. This adaptive rule ensures that each RL round halts once learning efficiency 
plateaus, while also transferring the model's hardest failure examples into the SFT stage, 
preventing unnecessary computation and reducing overfitting.

\subsubsection{Refining Model Weaknesses via SFT}\label{sec:sft}
This phase represents the SFT component of DPPO and serves as the complementary counterpart to the exploratory RL stage. While the RL phase reveals model weaknesses through exploratory interaction, the SFT phase refines the weaknesses exposed by RL, transforming exploratory insights into strengthened and generalized capabilities.

\vspace{4pt}
\noindent
\textbf{Data construction.}  
Following each RL round, all rollout trajectories are analyzed to identify both improved behaviors and unresolved weaknesses. Rather than relying solely on successful rollouts, we explicitly target the model’s unmastered abilities to construct a new supervision corpus for fine-tuning.  
The enhanced dataset \(\mathcal{D}_{\text{SFT}}\) is composed of three complementary sources:
$
\mathcal{D}_{\text{SFT}} = \mathcal{D}_{\text{weak}} \cup \mathcal{D}_{\text{rel}} \cup \mathcal{D}_{\text{gen}}$,
where \(\mathcal{D}_{\text{weak}}\) consists of hard or incorrectly solved rollout samples identified by the Difficulty-Aware Sampling mechanism in the RL phase, \(\mathcal{D}_{\text{rel}}\) includes related embodied samples retrieved from the dataset according to the weak ability dimensions, and \(\mathcal{D}_{\text{gen}}\) contains general data to replay and avoid the forgetting. Together, these components enable the SFT phase to perform targeted knowledge infusion, transforming the model’s observed weaknesses into structured supervision signals for continual policy reinforcement.

\vspace{4pt}
\noindent
\textbf{Refining exposed weaknesses.} After constructing \(\mathcal{D}_{\text{SFT}}\), the model undergoes SFT to consolidate the skills discovered during reinforcement learning. By globally integrating localized RL improvements into the policy distribution, the model mitigates catastrophic forgetting and achieves stronger generalization across diverse embodied domains.

\subsection{DPPO: Unified Preference Learning}\label{sec:expl}

The theoretical underpinning of our framework is the unification of SFT and RL into a cohesive paradigm of Preference Learning (PL). We suppose that these seemingly disparate training methodologies are specific instantiations of a single, universal objective:
\begin{equation}\label{eq:upl_main}
    \theta^* = \arg \max_{\theta} \mathbb{E}_{c\sim D_{pref}}[\log P(c|\pi_\theta)],
\end{equation}
The core difference lies in the specific structure of the preference sample $c$ and the probabilistic model $P(c|\pi_\theta)$. Specifically, for SFT, the preference sample $c$ is a single expert trajectory $\tau^*$. This simplifies the universal objective (Eq.~\ref{eq:upl_main}) into the standard, stable negative log-likelihood loss, $\mathcal{L}_{SFT}(\theta)$, which only considers positive exemplars:
\begin{equation}
    \log P(\tau^*|\pi_\theta) = \sum_{(c,a^*) \in \tau^*} \log \pi_\theta(a^*|c),
\label{eq:upl_sft}
\end{equation}

In contrast, for RL (i.e., GRPO), the sample $c$ is a ranked list of trajectories, (e.g., $\tau_i \succ \tau_j$). As shown in Appendix~\ref{sec:expl_sub}, this yields a comparative objective that learns from both superior and inferior examples. This unified formulation (Eq.~\ref{eq:upl_main}) reveals the precise \textbf{synergy of DPPO}. SFT (Eq.~\ref{eq:upl_sft}) acts as a direct and stable mechanism for knowledge enhancement, optimizing only on positive exemplars ($\tau^*$). RL (GRPO) performs weakness detection and refinement by learning from comparative samples ($c$) to correct subtle flaws that SFT alone overlooks. The DPPO Metaloop leverages this synergy to achieve robust, data-efficient skill acquisition. The full derivations are detailed in Appendix~\ref{sec:expl_sub}.

\section{Experiments}
\begin{figure*}[h!]
    \centering

    \setlength{\abovecaptionskip}{2pt}
    \setlength{\belowcaptionskip}{2pt}
    \setlength{\intextsep}{2pt}
    \setlength{\textfloatsep}{2pt}

    \setlength{\tabcolsep}{1pt}
    \setlength{\floatsep}{2pt}
    \setlength{\dblfloatsep}{2pt}

    \newcommand{\subfigvspace}{\vspace{-6pt}}  

    \begin{subfigure}[b]{0.32\linewidth}
        \centering
        \includegraphics[width=\linewidth]{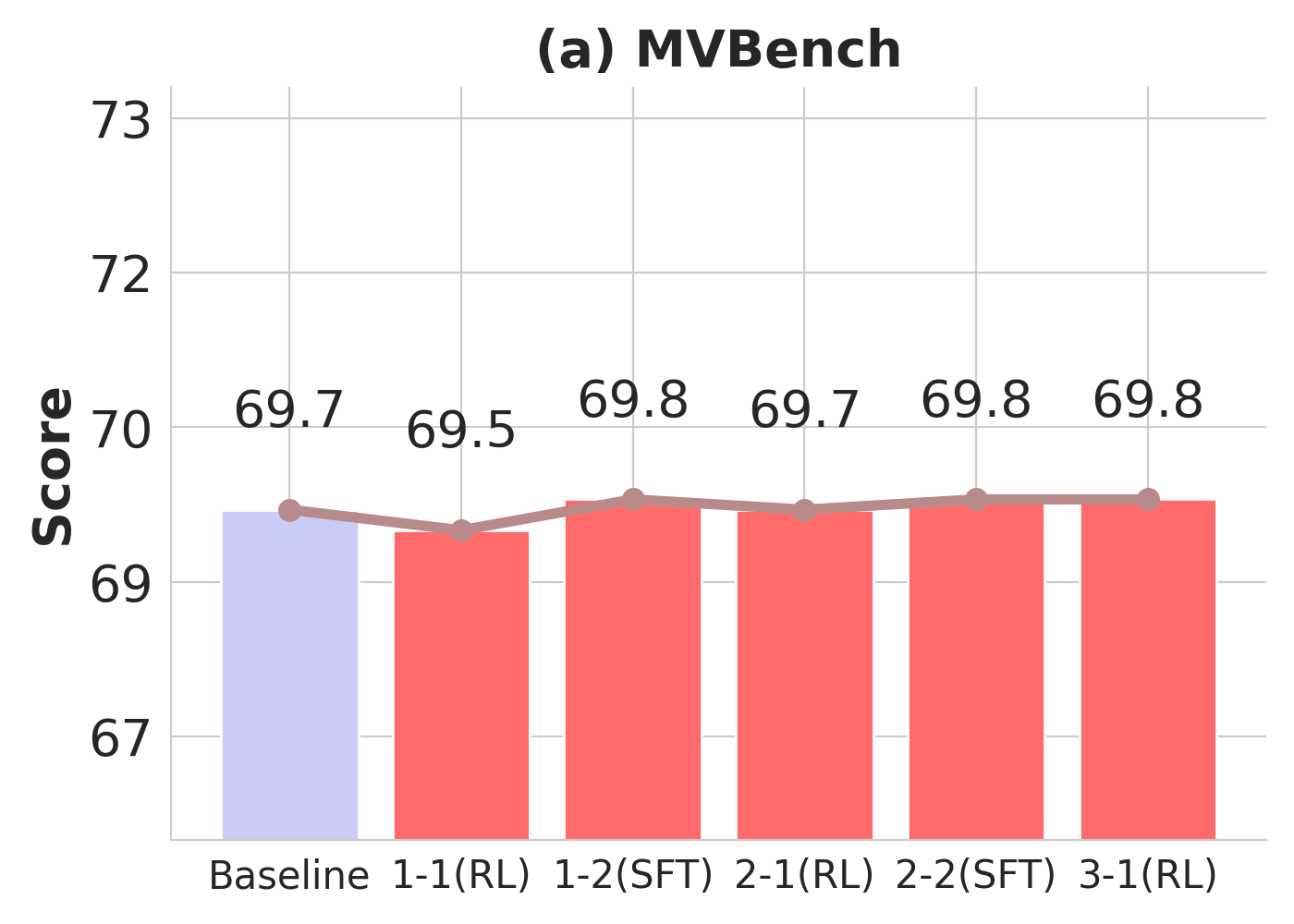}
    \end{subfigure}
    \hfill
    \begin{subfigure}[b]{0.32\linewidth}
        \centering
        \includegraphics[width=\linewidth]{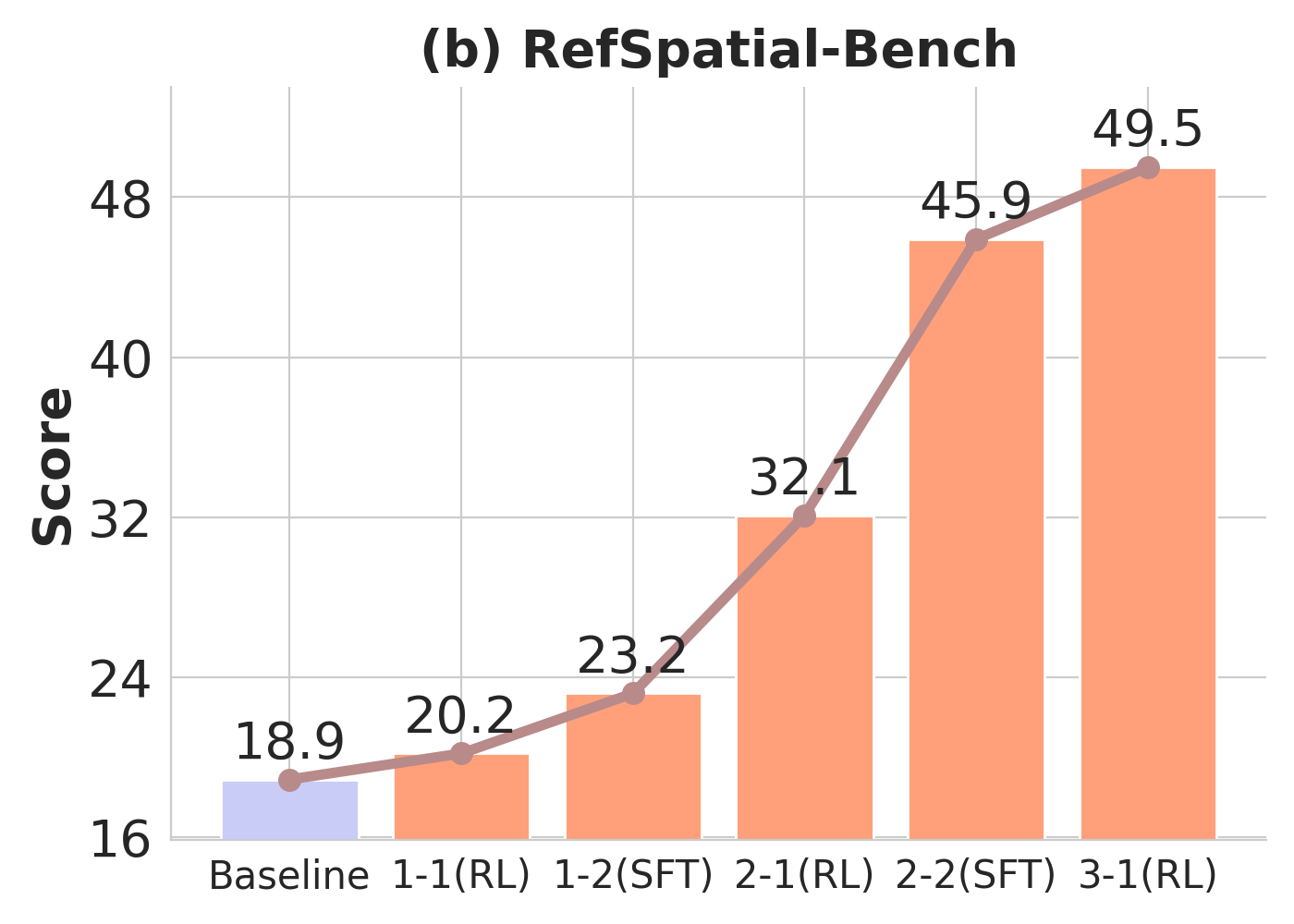}
    \end{subfigure}
    \hfill
    \begin{subfigure}[b]{0.32\linewidth}
        \centering
        \includegraphics[width=\linewidth]{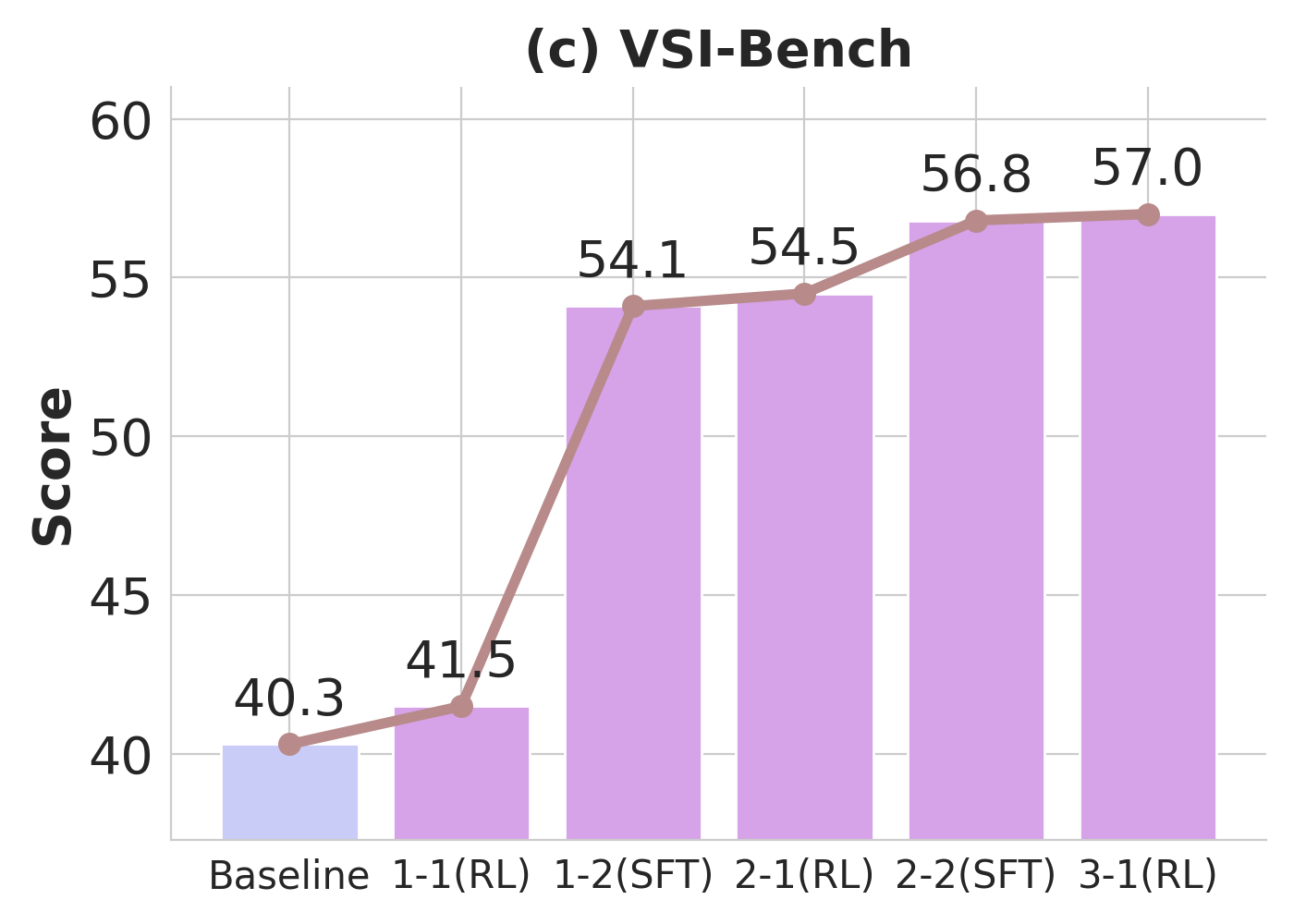}
    \end{subfigure}

    \subfigvspace  

    \begin{subfigure}[b]{0.32\linewidth}
        \centering
        \includegraphics[width=\linewidth]{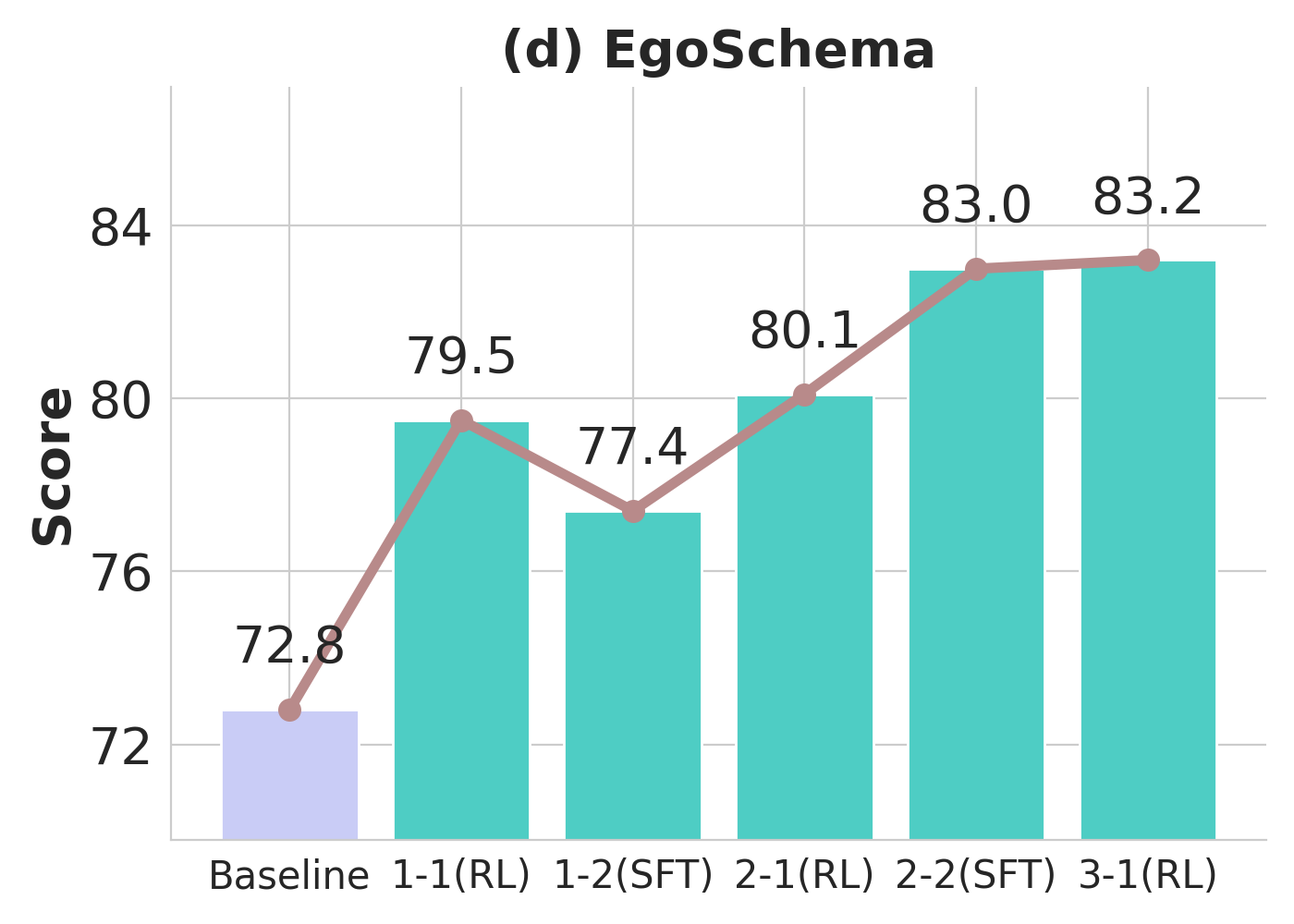}
    \end{subfigure}
    \hfill
    \begin{subfigure}[b]{0.32\linewidth}
        \centering
        \includegraphics[width=\linewidth]{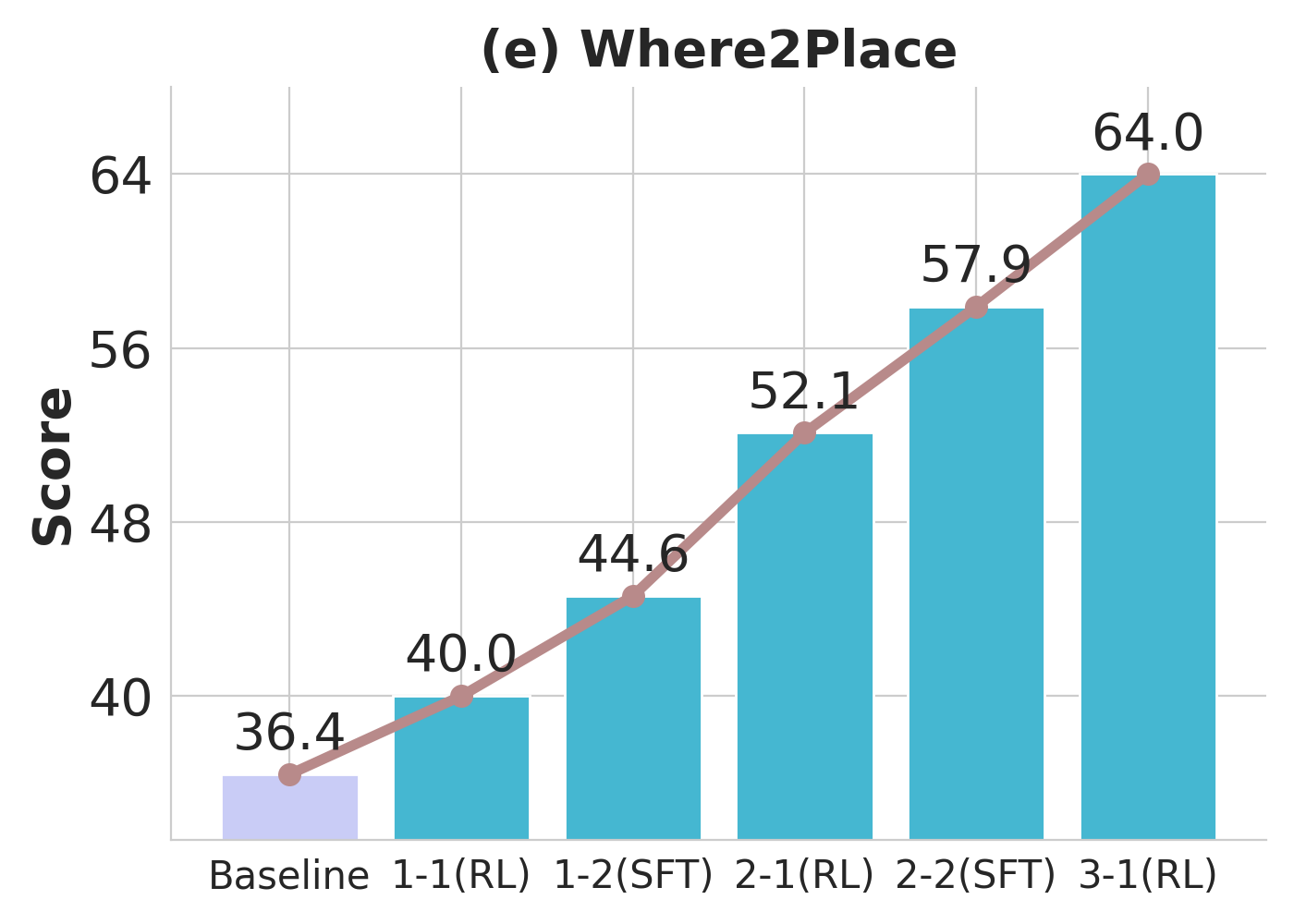}
    \end{subfigure}
    \hfill
    \begin{subfigure}[b]{0.32\linewidth}
        \centering
        \includegraphics[width=\linewidth]{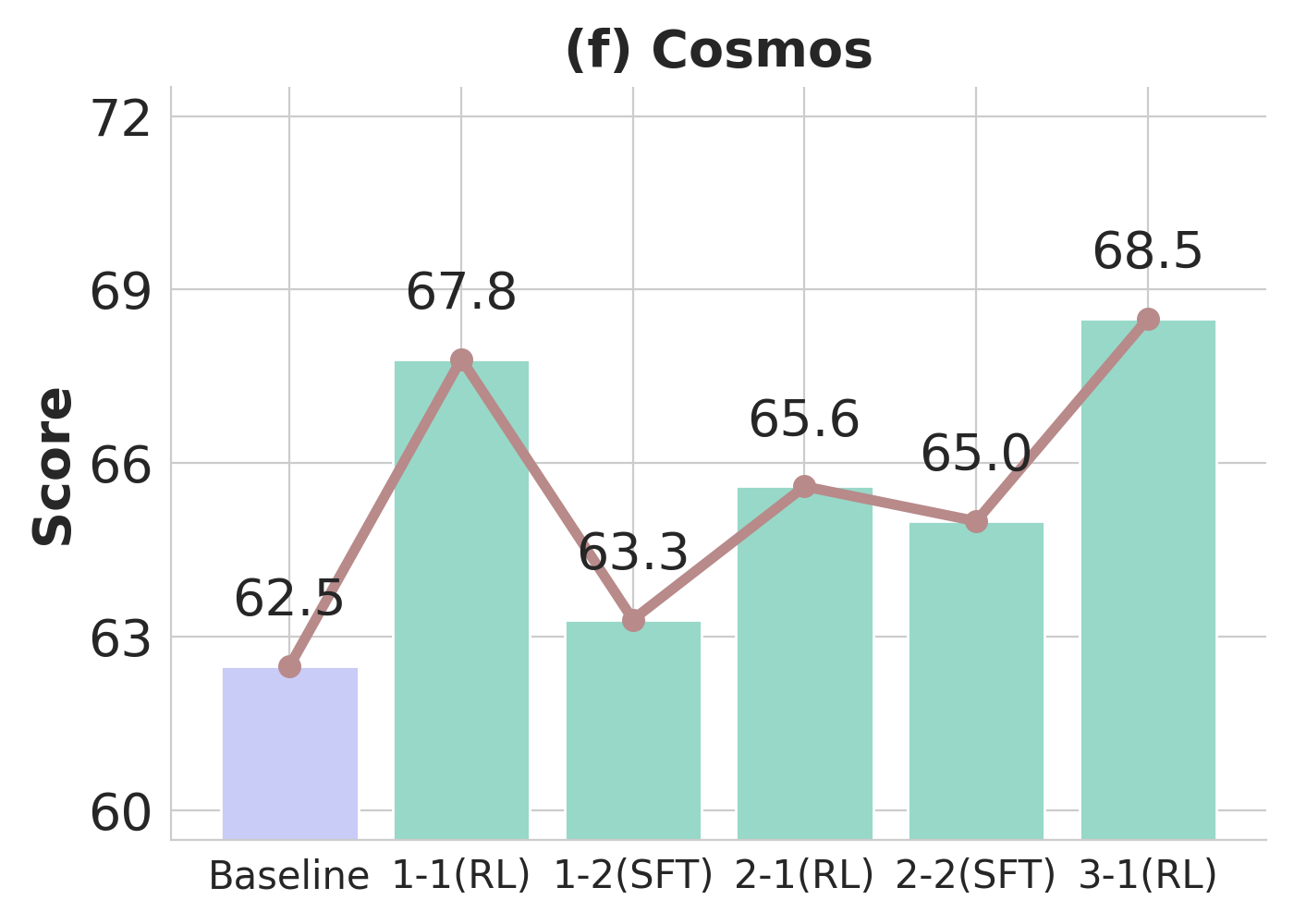}
    \end{subfigure}

    \caption{
        Performance evolution across the stages of Pelican-VL 72B.
        The model exhibits continuous improvements on embodied benchmarks while maintaining stable results on general datasets, 
        as demonstrated by its consistent performance on MVBench, a general-domain benchmark \textbf{(Finding~1)}.
    }
    \label{fig:evolution}
\end{figure*}

\label{sec:expriments}\subsection{Experimental Setup \label{sec:exp}}
\paragraph{Training data.}
{
For Pelican-VL 7B, we curate \textbf{200K} instances for SFT and \textbf{194K} for RL, 
categorizing the data into four fundamental capability areas to specifically address 
challenges in embodied AI, e.g., spatial perception, temporal reasoning, etc. 
More details can be viewed in Appendix~\ref{sec:data_suppl}.}

\vspace{4pt}
\noindent \textbf{Metaloop setting.}
We conduct three metaloop, each loop consisting of an RL phase followed by an SFT phase. We adopt a curated embodied data set filtered by temporal length to ensure compact interactions focused on action. In the first loop, training data are restricted to video segments shorter than 32\,s, allowing the model to focus on short-horizon manipulation and spatial reasoning tasks. In the second loop, the temporal limit is relaxed to 64\,s, enabling the model to explore longer and more compositional trajectories as its competence grows. Moreover, due to variations in task difficulty and data source, the frame extraction settings differ across datasets; each video clip is sampled up to 32 frames per episode. During RL training, each rollout sequence contains up to 16 times, forming the effective temporal context for policy optimization. This progressive expansion of temporal coverage across cycles encourages the policy to generalize from short, focused tasks to more extended, temporally dependent behaviors.

\definecolor{morandiRed}{RGB}{203,160,152}
\definecolor{morandiBlue}{RGB}{173,190,202}
\definecolor{morandiBorder}{RGB}{180,180,180} 

\newcolumntype{C}[1]{>{\centering\arraybackslash}p{#1}}

\begin{figure*}[h!]
    \centering

    \begin{subfigure}[b]{0.24\linewidth}
        \centering
        \includegraphics[width=\linewidth]{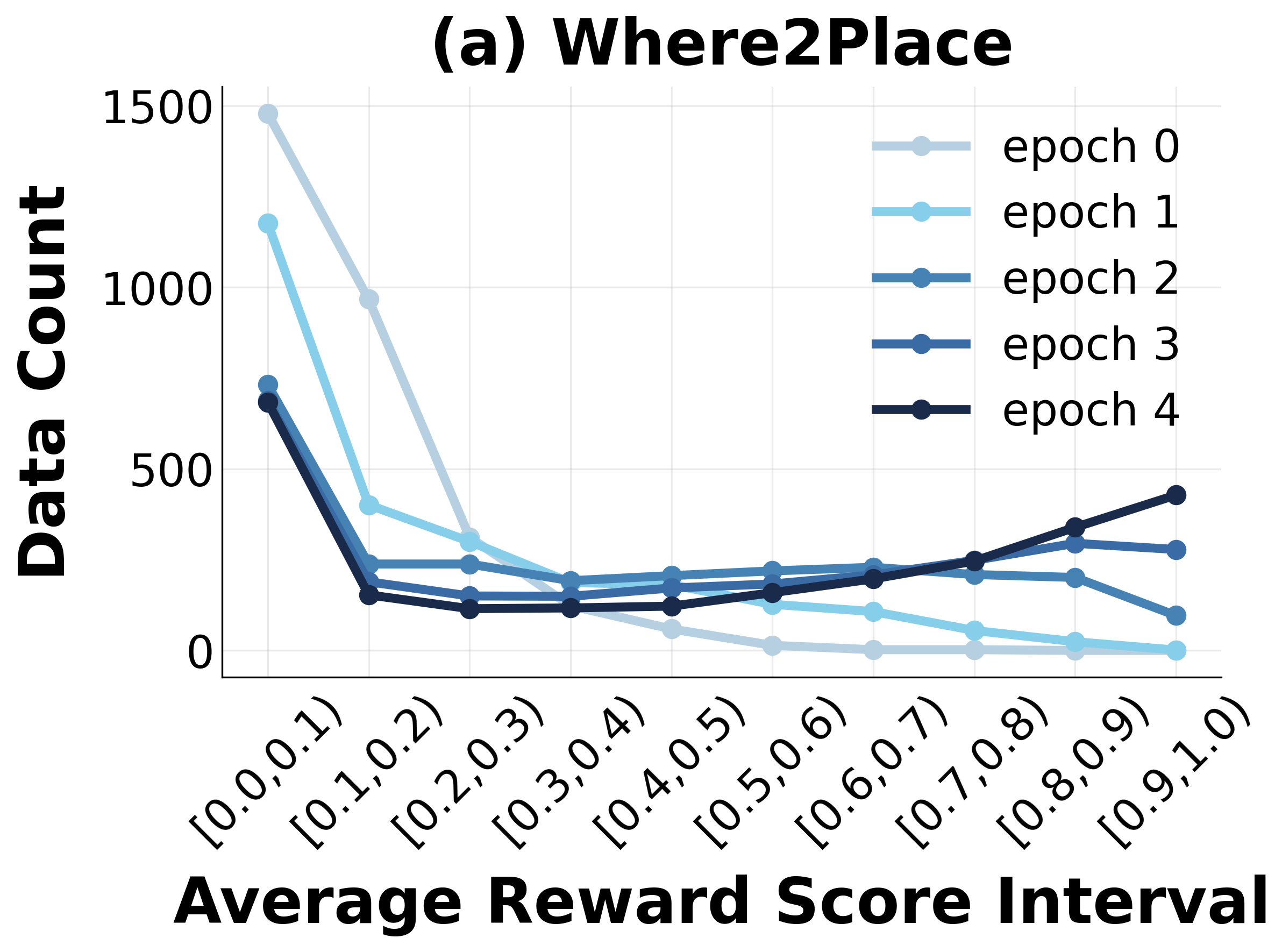}
    \end{subfigure}
    \begin{subfigure}[b]{0.24\linewidth}
        \centering
        \includegraphics[width=\linewidth]{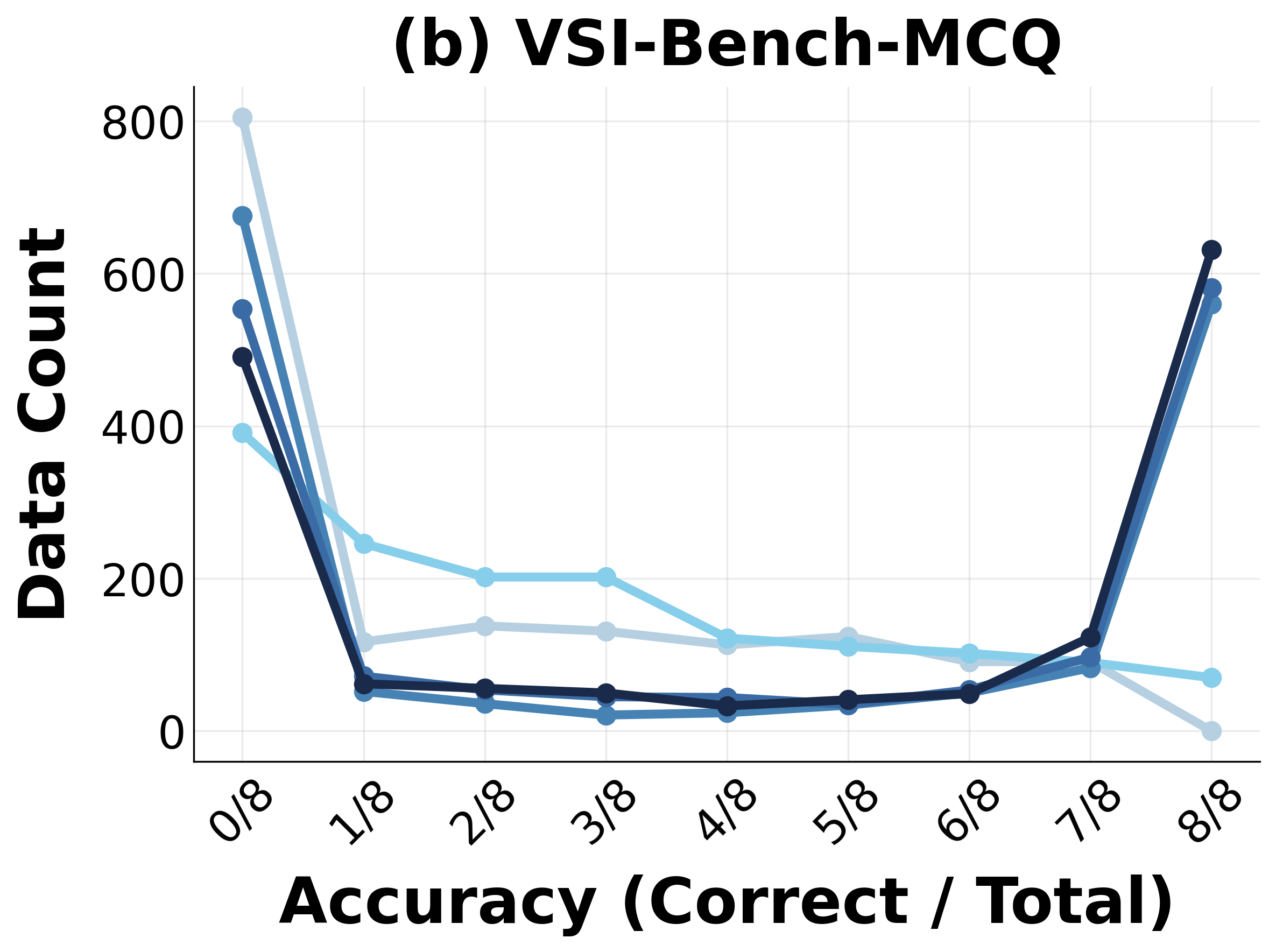}
    \end{subfigure}
    \begin{subfigure}[b]{0.24\linewidth}
        \centering
        \includegraphics[width=\linewidth]{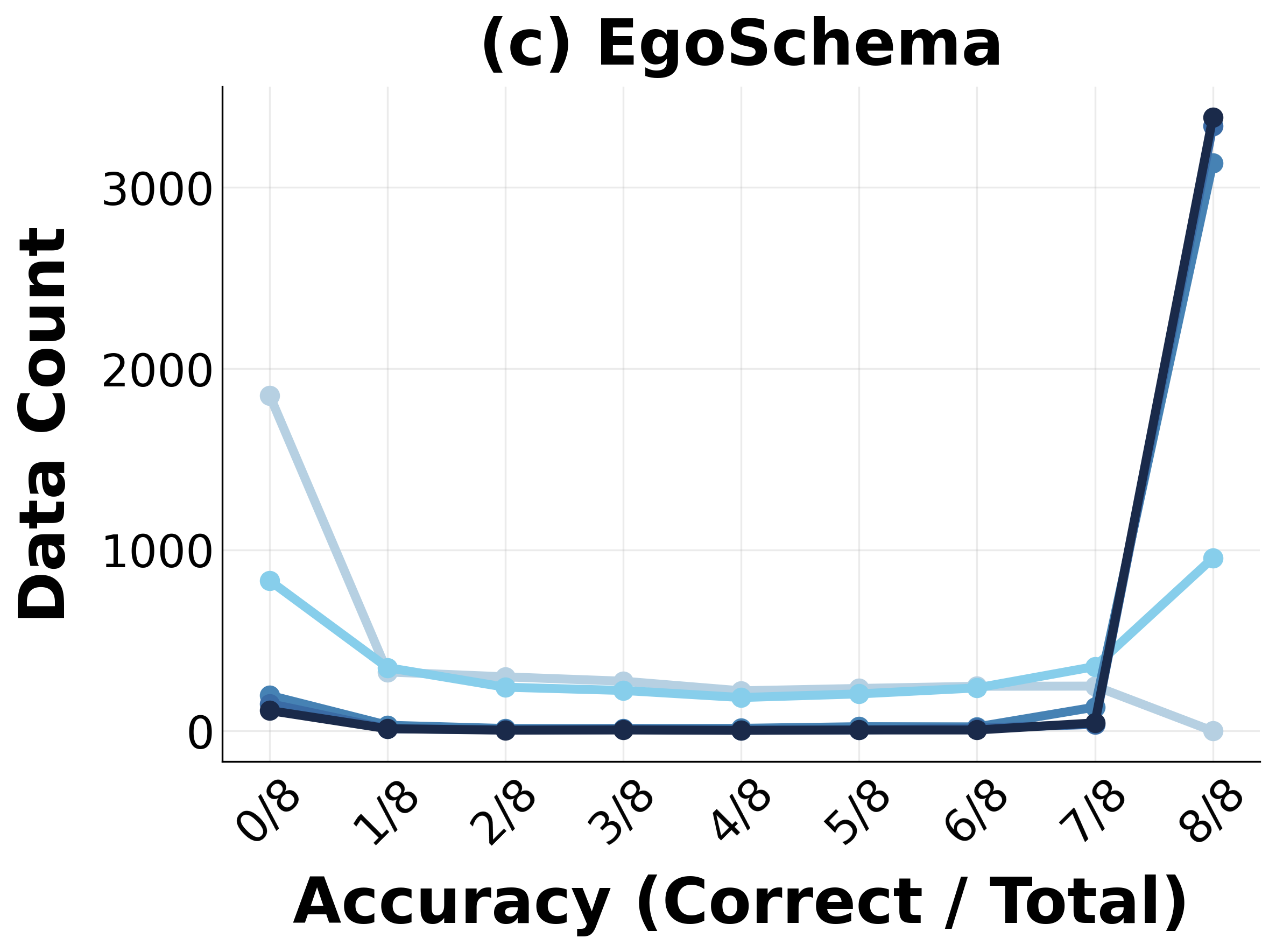}
    \end{subfigure}
    \begin{subfigure}[b]{0.24\linewidth}
        \centering
        \includegraphics[width=\linewidth]{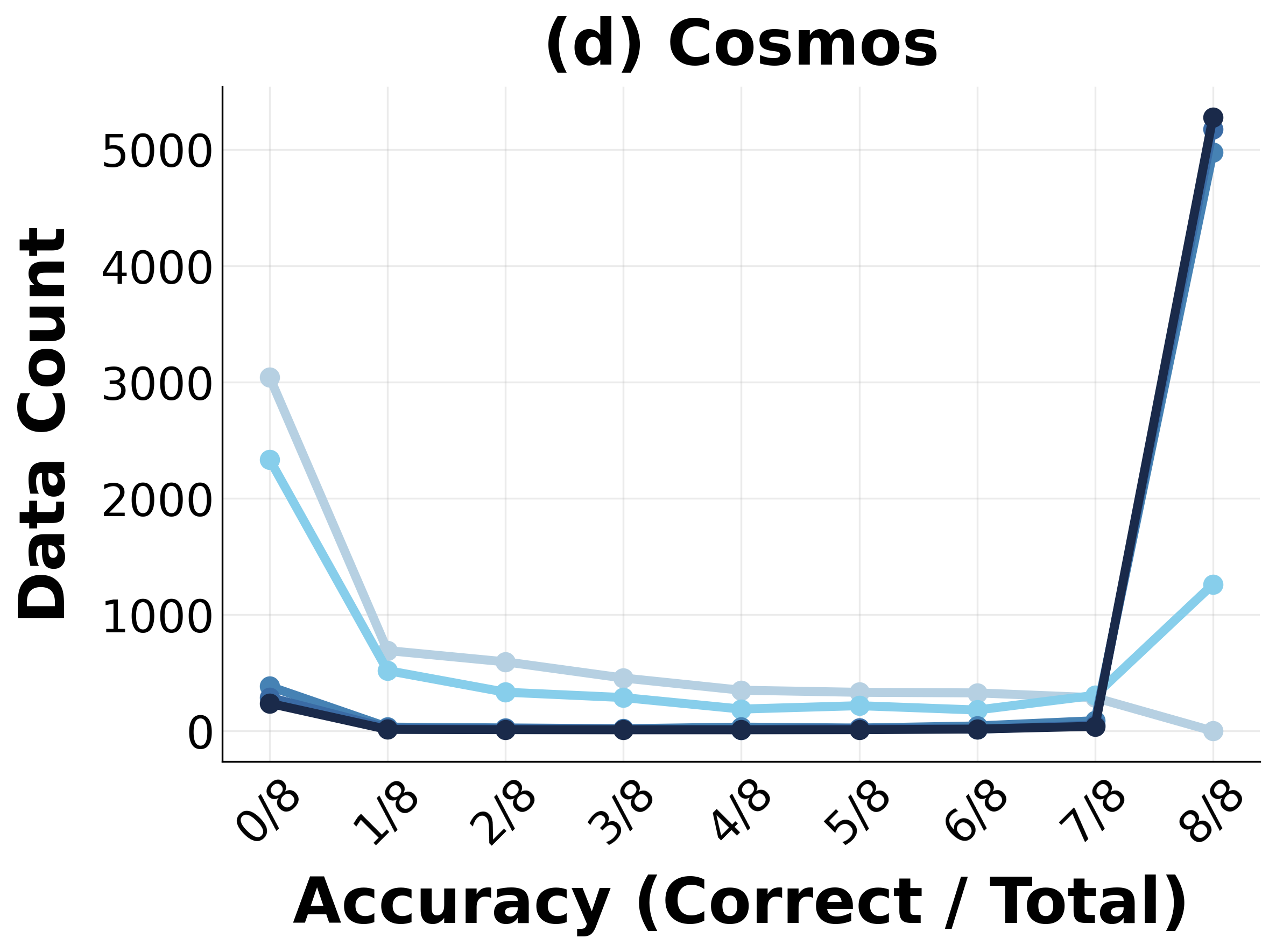}
    \end{subfigure}
    \caption{Distributional shift of training data relative to distinct benchmarks in (1-1) RL training on the Pelican-VL 7B model. 
For Where2Place, the rewards are numerical and model performance is measured by the average score per rollout. For the other datasets, the rewards are binary, and performance is measured by the number of correct answers. The progressively darker line colors indicate the progression of RL training, where we observe a steady reduction in unlearned tasks and a corresponding increase in successfully solved tasks. \textbf{(Finding~2)}.}
\label{performance}
\end{figure*}

\subsection{Performance Evolution Trajectory Across Metaloop Training \label{sec:cyclebound}}
To demonstrate the effectiveness of our metaloop approach, we divide the training process into three loops. Each loop consists of one RL phase and one SFT phase (i.e., 1-1 and 1-2 in Fig.~\ref{fig:evolution}), with the process concluding with a final RL phase (3-1). We conduct experiments on several benchmarks which including MVBench~\cite{mvbench}, EgoSchema~\cite{EgoSchema}), (RefSpatialBench~\cite{song2025robospatial}, VSI-Bench~\cite{yang2024think}, Where2Place~\cite{yuan2024robopoint}, and COSMOS~\cite{seghal2023cosmos}. From Fig.~\ref{fig:evolution}, we observe that \ul{the performance of 72B size Pelican on the other five embodied benchmarks progressively improves with each training loop.} Furthermore, \ul{the additional performance gains in COT are attributed to the final RL phase}, as demonstrated in Tab.~\ref{tab:7b_training_log_single_col} and Tab.~\ref{tab:7b_training_cot} (Appendix~\ref{sec:exp_suppl}).

To further examine the fine-grained performance of the model, we conduct an analysis during the RL stage, focusing on changes in the distribution of training data, specifically, the shifts in the distribution of questions the model can and cannot answer. Fig.~\ref{performance} shows the evolution of the 7B model's data distribution on benchmark-related tasks. We observe that \ul{as model training progresses, the number of questions the policy model is unable to answer gradually decreases, while the number it can answer continues to increase.} For example, in VSI-Bench-MCQ, the number of items for which the model gets 0 out of 8 correct decreases from 800 to 500, indicating that the number of questions the model cannot answer at all has been reduced.

\begin{figure}[htbp]
    \centering
    \includegraphics[width=0.6\linewidth,keepaspectratio]{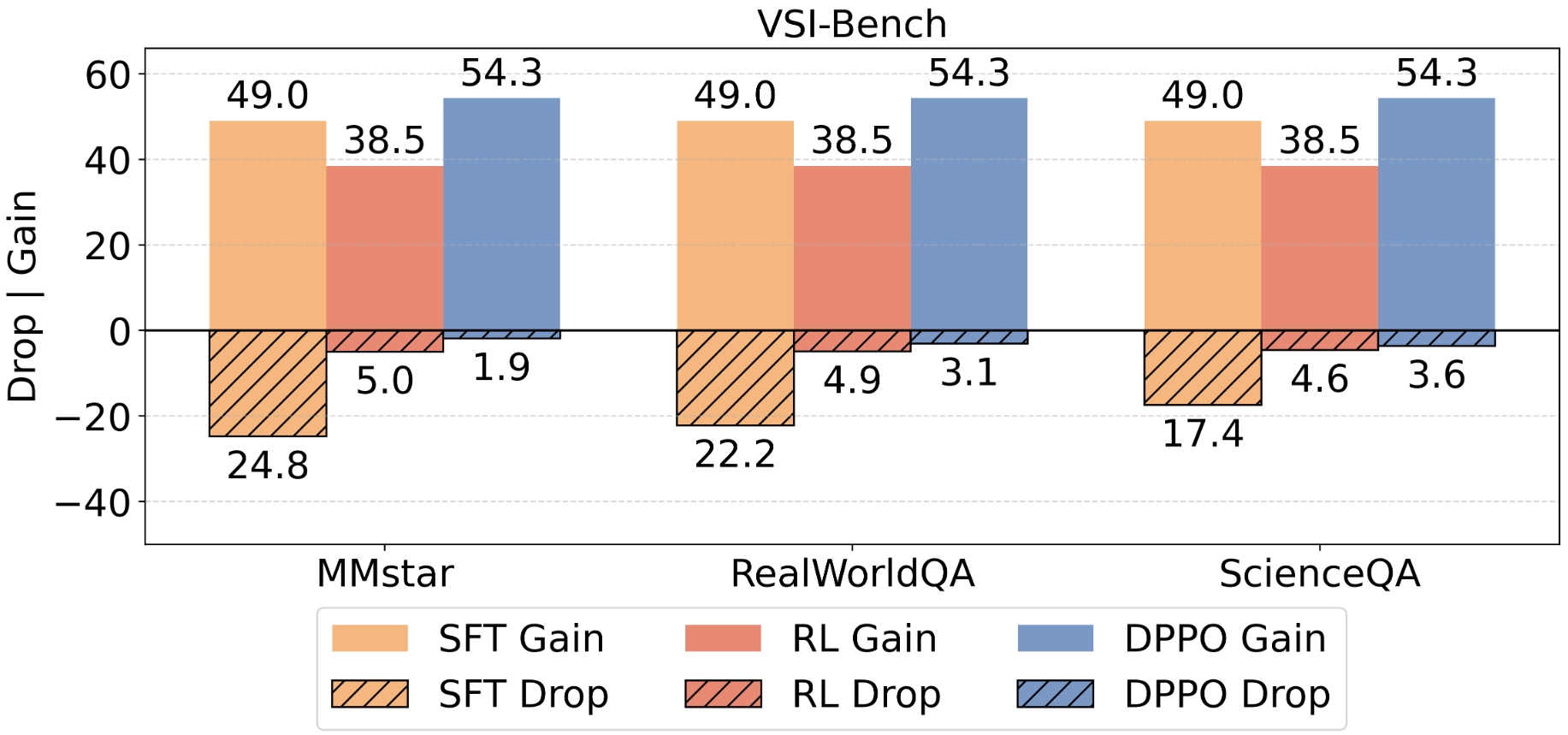}
    \caption{Comparison of SFT, RL, and DPPO on the 7B model in terms of performance gain on the VSI-Bench and forgetting on general benchmarks. The results demonstrate a substantial performance gain of 54.3, while the observed performance degradation remains notably limited, for example, 1.9 for DPPO, 5.0 for SFT, and 24.8 for RL on the MMStar dataset \textbf{(Finding~3)}.}
    \label{fig:CatastrophicForgetting}
\end{figure}

\paragraph{Catastrophic forgetting.} MVBench is a general dataset used to measure the model’s performance in the general domains. The model’s performance on MVBench can indicate whether forgetting occurs during training. Fig.~\ref{fig:evolution} (a) shows that the model’s performance remains stable throughout training, meaning no obvious forgetting occurs during the metaloop training. To gain deeper insights into the phenomenon of forgetting, we train models on embodied data and evaluate the performance of various approaches (SFT, RL, and DPPO) across multiple testsets (MMstar, RealWorldQA, and ScienceQA). As Fig.~\ref{fig:CatastrophicForgetting} depicts, \ul{DPPO achieves greater performance gains(+15.8\% over RL), while exhibiting less performance drop on unseen datasets.} 
\begin{figure*}
    \centering
    \includegraphics[width=1\linewidth]{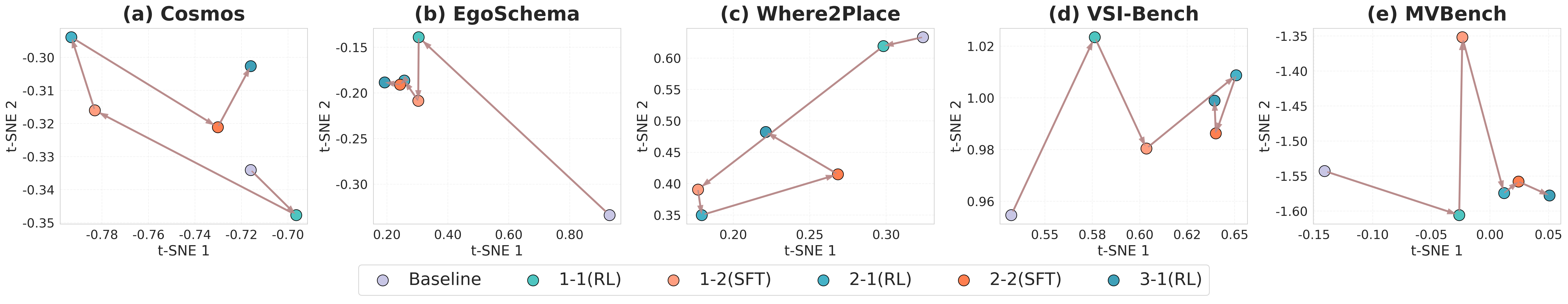}
    \caption{Distributional shift of the Pelican-VL 7B model's trajectory embedding centroids across DPPO metaloop cycles, visualized via t-SNE on five benchmarks. The figure tracks how the model's representations evolve over alternating training steps, producing distinct trajectories for each benchmark. Visualization details are provided in Sec.~\ref{part:distshift}. The divergent trajectories across benchmarks illustrate that different tasks shift the model in different representational directions, making a single training stage insufficient and motivating our multi-stage DPPO metaloop design \textbf{(Finding~5)}.}
    \label{fig:stage_clusters_comparison}
\end{figure*}

\paragraph{Advantages of DPPO compared with SFT or RL.} 
\begin{table}[htbp]
\centering

\setlength{\abovecaptionskip}{2pt}
\setlength{\belowcaptionskip}{2pt}

\caption{
Performance comparison with adopting only SFT and RL methods under the same data budget for Pelican-VL 7B. 
DPPO consistently delivers stronger overall performance and markedly better generalization on diverse benchmarks.
\textbf{(Finding~4)}.
}
\label{tab:7b_training_log_no_cycle}

\small
\renewcommand{\arraystretch}{0.95}   
\setlength{\tabcolsep}{2pt}          

\begin{tabular}{lccccccc}
\toprule
\textbf{Model} & \textbf{Ego.} & \textbf{W2P} & \textbf{Omni.} & 
\textbf{RefS.} & \textbf{VSI-B.} & \textbf{Average}\\
\midrule
Base & 59.6 & 24.0 & 39.2 & 7.4 & 37.2 & 33.5 \\
\midrule
RL   & 72.0 & 27.8 & 45.3 & 12.2 & 46.1 & 40.7 \\
SFT  & 57.1 & 32.0 & 42.9 & 14.1 & 53.8 & 39.9 \\
\midrule
Our  & \textbf{72.5} & \textbf{53.5} & \textbf{46.5} & 
\textbf{28.2} & \textbf{54.3} & \textbf{51.0}\\
\bottomrule
\end{tabular}
\end{table}

 After applying difficulty-aware sampling, the performance on datasets such as EgoSchema and Cosmos improves rapidly, whereas VSI and Where2Place exhibit noticeably slower learning curves, as shown in Fig.~\ref{performance}.
This highlights a challenging exploration landscape in which standard training approaches fail to learn efficiently. 
DPPO addresses this through an iterative metaloop: an RL phase with progressive task-difficulty shaping enables the model to acquire stable and learnable patterns without collapsing, followed by a targeted SFT phase that consolidates underperforming cases while mitigating the semantic overfitting commonly observed in pure SFT.


The performance trends in Tab.~\ref{tab:7b_training_log_no_cycle} and 
Fig.~\ref{fig:evolution} demonstrate that our DPPO algorithm benefits substantially 
from the iterative metaloop: \ul{it preserves capability balance, mitigates catastrophic 
forgetting, and consistently elevates task-specific skills, ultimately achieving 
significantly stronger generalization than either pure SFT or standalone RL. }
Notably, while we find that the performance on some datasets does not continue to 
increase by stage~3, we still observe a pronounced improvement in the model’s 
chain-of-thought reasoning and deliberation behaviors. 
This suggests that stage~3 primarily strengthens deeper generalization and reasoning 
abilities that extend beyond task-level accuracy, as further analyzed in 
Appendix~\ref{sec:cot_comparison}.



\subsection{Final Result}\label{sec:finalresult}
\paragraph{Cycle distribution shift.}
\label{part:distshift}
To visualize the distribution shift across RL and SFT stages, we extract trajectories from each benchmark at every stage checkpoint. We then obtain trajectory embeddings using Qwen3-Embedding-8B \cite{zhang2025qwen3}, reduce their dimensionality to 2D with t-SNE, and compute the centroid of each stage’s embedding cluster via K-Means.
In Fig.~\ref{fig:stage_clusters_comparison}, a clear distribution shift is observed after the first round of RL and subsequent SFT across the three embodied benchmarks: Cosmos, VSI-Bench, and Where2Place. This indicates \ul{a capability gap between the embodied model and the base model, suggesting that the embodied tasks are unfamiliar to the base model and require targeted learning.} In contrast, during the second round of RL and SFT, the distribution shift becomes smaller and remains close to the position of the first SFT stage, which aligns with the results shown in Fig.~\ref{fig:evolution}, where performance changes in the second RL–SFT cycle are relatively minor compared to the first.

\vspace{4pt} \noindent \textbf{Benchmark performance.} Tab.~\ref{tab:main_performance} presents a performance comparison between Pelican-VL and other baseline models on multiple benchmarks. As shown in Tab.~\ref{tab:100b_models}, \ul{Pelican-VL 1.0 achieves superior performance at the 100B-level. Compared with the Qwen2.5-VL 72B-Instruct, Pelican-VL 1.0 achieves significant improvements on benchmarks in the embodied domain.} This is mainly attributed to our training framework, where RL can accurately identify currently deficient capabilities, and SFT is employed to inject knowledge for weak capabilities, expanding the model's capability boundary. Furthermore, since we incorporated general data in the second round of metaloop, compared with the base model, we observe that while enhancing the model's embodied capabilities, its general capabilities do not experience a significant degradation.

\begin{table*}
\centering
\caption{Overall performance comparison on benchmarks. Bold and underlined numbers indicate the best and second-best results, respectively. A dagger ($\dagger$) marks results differing from official reports, possibly as official evaluations used model-specific prompts and the models are prompt-sensitive, while our results are obtained under a unified protocol for fair comparison. An asterisk (*) denotes results reported from official sources. We can observe that our model achieves strong performance under different benchmarks \textbf{(Finding~6)}.}
\label{tab:main_performance}

\begin{subtable}{\textwidth}
    \centering
    \caption{Performance of Models with $\leq$100B Parameters.}
    
    \label{tab:100b_models}
    \resizebox{\textwidth}{!}{%
    
    \begin{tabular}{llcccccccc >{\columncolor{pelicanlight}}c >{\columncolor{pelicanmain}}c}

    \toprule
    Category & Benchmark & Qwen2.5-VL & InternVL3.5 & Qwen2.5-VL & Qwen3-VL & Qwen3-VL & InternVL3.5 & InternVL3.5 & Qwen2.5-VL & Pelican-VL & Pelican-VL \\
    & & 7B-Instruct & 8B & 32B-Instruct & 30B-A3B-Instruct& 30B-A3B-Thinking & 30B-A3B & 38B & 72B-Instruct & 7B & 72B \\
    \midrule
    Common & MVBench & 68.1 & \underline{73.5} & 66.4 & 71.6 & 71.7 & 72.1* & \textbf{76.2} & 69.7 & 67.7 & 69.7 \\
    \midrule
    \multirow{8}{*}{Spatial-Physical}& RoboSpatial & 44.3 & 48.0 & 48.6 & 47.4$^{\dagger}$ & 57.4 & 44.3 & 56.3 & 47.7 & \underline{57.5} & \textbf{61.1} \\
    & BLINK & 55.9 & 59.0 & 62.6 & 62.3 & \textbf{67.0} & \underline{63.1} & 59.3 & 62.1 & 56.8 & 60.3 \\
    & PhyX & 43.4 & 39.1 & 62.3 & 76.7 & \underline{84.4} & 53.4 & 39.1 & 53.1 & 80.1 & \textbf{86.4} \\
    & OmniSpatial & 40.1 & 44.6 & 37.1 & 48.5 & \textbf{50.7} & 46.4 & 49.0 & 48.7 & 43.5 & \underline{49.6} \\
    & Where2Place & 25.8 & 32.3 & 21.6 & 1.7$^{\dagger}$ & 36.6 & 28.3 & 36.1 & 38.1 & \underline{57.3} & \textbf{64.0} \\
     & EgoSchema & 57.5 & 61.2 & 66.5 & 71.4 & 68.2 & 70.5 & 69.0 & 70.9 & \underline{73.3} & \textbf{79.3} \\
    & EmbSpatiaBench & 71.1 & 74.4 & 74.9 & 75.6 & \textbf{79.2} & \underline{77.1} & 70.3 & 73.8 & 73.2 & 76.6 \\
    & RefSpatialBench & 16.7 & 23.2 & 16.4 & 0.6$^{\dagger}$ & 19.0$^{\dagger}$ & 14.3 & \underline{29.9} & 24.7 & 22.3 & \textbf{49.5} \\
    \midrule
    \multirow{3}{*}{CoT-Reasoning} 
    & ERQA & 40.0 & 39.3 & 43.5 & 43.3 & 43.0 & 43.0 & \textbf{44.7} & \underline{43.8} & 39.8 & 43.0 \\
    & COSMOS & 54.2 & 49.9 & 53.8 & 55.8 & 58.3 & 50.9 & 56.2 & \underline{62.5} & 60.2 & \textbf{68.5} \\
    & VSI-Bench & 37.3 & 56.0 & 38.3 & \underline{64.0} & 57.2 & \textbf{64.2} & 61.3 & 40.3 & 52.8 & 57.3 \\
    \midrule
    Average & & 46.2 & 50.0 & 49.3 & 51.6 & \underline{57.7} & 52.3 & 53.9 & 53.0 & 57.0 & \textbf{63.8} \\
    \bottomrule
    \end{tabular}%
    }
\end{subtable}

\vspace{0.2cm} 

\begin{subtable}{\textwidth}
    \centering
    \caption{Performance of Models with $>$ 100B Parameters.}
    \label{tab:200b_models}
    \resizebox{\textwidth}{!}{%
    \begin{tabular}{llcccccccc>{\columncolor{pelicanmain}}c}
    \toprule
    Category & Benchmark & Qwen3-VL & Qwen3-VL & InternVL3.5 & Qwen3-VL-Plus & GPT-5 & GPT-5-Mini & Gemini2.5-Flash & GPT-4o & Pelican-VL \\
    & & 235B-A22B-Thinking & 235B-A22B-Instruct & 241B-A28B & proprietary & proprietary & proprietary & proprietary & proprietary & 72B \\
    \midrule
    Common & MVBench & 74.5 & 76.1 & \textbf{78.4} & \underline{76.4} & 73.1 & 66.9 & 65.2 & 64.4 & 69.7 \\
    \midrule
    \multirow{8}{*}{Spatial-Physical} & RoboSpatial & \textbf{62.0} & 58.0 & 51.1 & 59.1 & 53.4 & 50.6 & 59.9 & 46.9 & \underline{61.1} \\
    & BLINK & 67.4 & 68.6 & 61.4* & \underline{69.3} & \textbf{69.9} & 67.4 & 62.6 & 64.2 & 60.3 \\
    & PhyX & \underline{85.3} & 63.2 & 62.6 & 63.9 & 83.6 & 82.5 & 77.7 & 41.2 & \textbf{86.4} \\
    & OmniSpatial & 53.8 & 52.1 & 53.2 & 52.9 & \textbf{57.6} & \underline{56.2} & 43.6 & 41.7 & 49.6 \\
    & Where2Place & \underline{52.2} & 40.0 & 35.1 & 46.3 & 38.6 & 31.5 & 35.1 & 20.3 & \textbf{64.0} \\
    & EgoSchema & 72.1 & 77.7 & 72.9 & \underline{78.0} & 73.7 & 68.8 & 61.8 & 69.2 & \textbf{79.3} \\
    & EmbSpatialBench & \textbf{83.4} & 82.4 & 79.1 & \underline{82.6} & 82.3 & 79.3 & 74.6 & 71.9 & 76.6 \\
    & RefSpatialBench & \underline{39.4}$^{\dagger}$ & 20.6$^{\dagger}$ & 23.6 & 19.9 & 21.6 & 13.9 & 35.4 & 9.9 & \textbf{49.5} \\
    \midrule
    \multirow{3}{*}{CoT-Reasoning} 
    & ERQA & 48.0 & 47.5 & 47.3 & 48.0 & \textbf{60.0} & \underline{53.5} & 49.8 & 37.8 & 43.0 \\
    & COSMOS & 62.3 & 60.4 & 56.4 & 53.0 & \underline{64.8} & 61.8 & 30.3 & 52.4 & \textbf{68.5} \\
    & VSI-Bench & 60.9 & \underline{62.6}* & \textbf{67.0} & 61.2 & 56.2 & 48.2 & 46.4 & 34.0 & 57.3 \\
    \midrule
    Average & & \underline{63.4} & 59.1 & 57.4 & 59.2 & 61.2 & 56.7 & 53.5 & 46.2 & \textbf{63.8} \\
    \bottomrule
    \end{tabular}%
    }
\end{subtable}

\end{table*}

Tab.~\ref{tab:200b_models} shows that Pelican-VL 1.0, trained solely on a diverse dataset consisting of 1M trajectories and 100K objects, demonstrates superior performance over even the top-performing 200B-level closed-source models, including GPT-5 and Gemini2.5-Flash. This further validates the effectiveness of our training strategy.
\subsection{Fine-Grained Capability Analysis}

We conducted a fine-grained capability analysis because current embodied benchmarks are ``black-box'' and too coarse-grained. They provide only task-level Pass/Fail metrics, which is insufficient for diagnostic analysis and cannot guide efficient, iterative model updates. To address this, we first defined a new 9-dimensional capability taxonomy (e.g., Physical \& Causal Reasoning, Decision \& Task Planning) and re-annotated 27,667 samples from public datasets to create a new diagnostic benchmark (see Appendix~\ref{sec:bench_suppl} for full taxonomy and dataset analysis).

Our analysis reveals two critical conclusions: (1) \ul{Existing benchmarks are severely imbalanced, with critical, real-world abilities being ``severely underrepresented''}, such as Physical \& Causal Reasoning (3.5\%) and Decision \& Task Planning (3.3\%). (2) Our Pelican-VL 1.0, trained with DPPO, demonstrates its most pronounced enhancements precisely in these difficult, neglected dimensions, achieving SOTA performance (e.g., a 15.4\% average uplift). This validates that \ul{our DPPO framework is uniquely data-efficient and highly effective at targeted capability refinement.}

\section{Discussions}
\label{sec:discussion}

In this work, we introduce Pelican-VL 1.0, to our knowledge, the largest open-source embodied brain model to date, along with DPPO, a novel, capital-efficient metacognitive framework used to train it. Our results demonstrate the framework’s exceptional effectiveness and achieve a 20.3\% performance uplift from its base model, outperforming 100B-level open-source counterparts by 10.6\%. By open-sourcing our complete models and reproducible toolchain, we aim to provide a foundation that breaks the resource bottleneck for the community. Pelican-VL 1.0 and DPPO mark not the end, but rather the first step toward our ultimate vision of autonomous, \texttt{self-evolving} intelligence. Our future work will focus on building a comprehensive self-evolving ecosystem, integrating autonomous DPPO, diagnostic benchmarks, capability-aligned data, and a closed hardware loop, to address key societal challenges and catalyze a transformative \textbf{ChatGPT moment for robotics}.

\newpage
\bibliography{reference.bib}
\bibliographystyle{plain}

\clearpage
\setcounter{page}{1}
\appendix

\section{Reward Detail}\label{sec:method_suppl}
The detailed reward definitions are summarized in Table~\ref{tab:reward_design}.
\begin{table*}
\centering
\small
\setlength{\tabcolsep}{6pt}
\renewcommand{\arraystretch}{1.35}
\caption{Rule-based multi-task reward design.}
\begin{tabular}{>{\centering\arraybackslash}m{3cm} | >{\centering\arraybackslash}m{3cm} | >{\centering\arraybackslash}m{9.5cm}}
\toprule
\textbf{Task Type} & \textbf{Reward Type} & \textbf{Description} \\
\midrule
Affordance Reasoning &
Affordance Reward &
Evaluates whether predicted manipulations are physically feasible and contextually valid, including grasping and placement correctness. \\
\hline
Counting and Distance Estimation &
Numeric Reward &
Rewards accurate quantitative perception and spatial measurement across objects and scenes. \\
\hline
Causal and Temporal Reasoning &
\multirow{4}{*}{Choice Reward} &
Assesses logical and temporal consistency in sequential task execution and causal dependencies. \\
\cline{1-1}\cline{3-3}
Task Success Evaluation & &
Determines whether the trajectory fulfills task-specific completion rules, such as target achievement or valid function-call execution. \\
\cline{1-1}\cline{3-3}
Task Planning & &
Rewards coherent hierarchical action decomposition that aligns sub-goals with overall objectives. \\
\cline{1-1}\cline{3-3}
Task Prediction & &
Measures accuracy in predicting task intents, expected outcomes, or next-step actions from multimodal inputs. \\
\bottomrule
\end{tabular}
\label{tab:reward_design}
\end{table*}

\begin{figure*}
    \centering
    \includegraphics[width=0.85\textwidth]{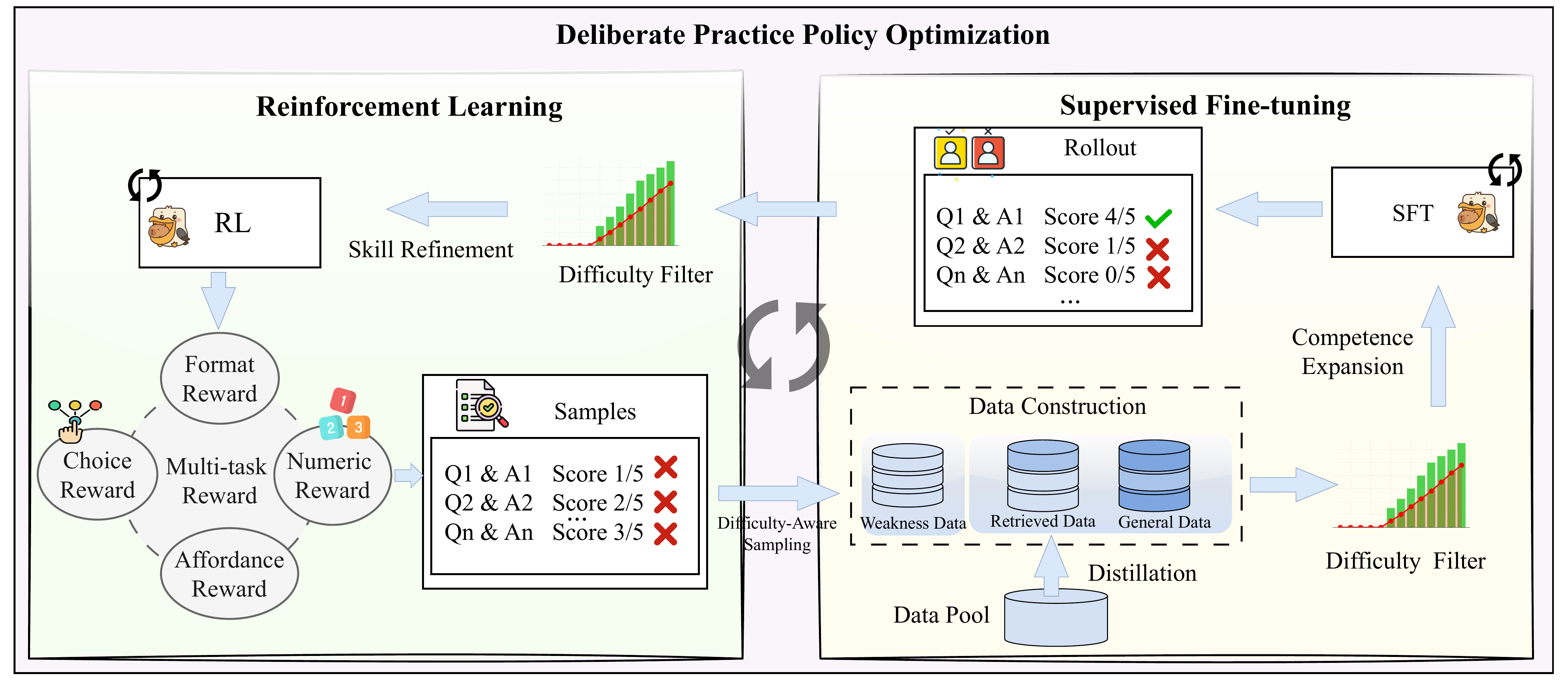}
    \caption{Overview of our training framework. This framework implements an iterative RL-SFT loop that leverages Rollout Logging and Difficulty-Aware Sampling to dynamically curate data. This adaptive data selection process is designed to achieve two complementary objectives: rapid capability enhancement during the RL phase and stable modal alignment during the SFT phase.}
    \label{fig:weakness}
\end{figure*}

\section{Data curation}\label{sec:data_suppl}
\subsection{Data for Training}
To advance the capability of Pelican-VL in perception, reasoning, and planning in real-world scenarios grounded in physical knowledge, we built a diverse data pool. This pool contains \textbf{231M} images and \textbf{29k} hours of video, encompassing \textbf{231M} open-ended question-answer (QA) pairs, \textbf{9M} grounding annotations (bounding boxes or keypoints), and \textbf{2M} multiple-choice questions (MCQs). From this curated data, we sampled \textbf{1.3M} instances for SFT stages and \textbf{0.5M} for RL, amounting to 4B training tokens,  and categorizing the data into four fundamental capability areas to specifically address challenges in embodied AI:

\begin{itemize}
    \item \textbf{Physical, Spatial and Numerical Reasoning.} This category aims to build a foundational understanding of the physical world. It includes data that assesses physics-grounded reasoning in visual scenarios (e.g., \textit{PhyX}, which contains multimodal questions across core physics domains) and the understanding of spatial configurations (e.g., \textit{VSI-bench}, with tasks on measurement estimation and spatiotemporal reasoning). We also integrate data from 2D images, 3D embodied videos, and simulated scenes to cover tasks like Reasoning QA (e.g., \textit{RefSpatial}).
    
    \item \textbf{Perception, Grounding and Multi-Object Consistency.} This category focuses on connecting language to visual elements and understanding object relationships. It leverages data that combines object and space-referenced information with VQA and detection. For instance, some datasets guide actions via image keypoint prediction from language instructions (e.g., \textit{Where2Place}), while others include referring expression tasks to ensure precise object grounding (e.g., \textit{RefSpatial}).
    
    \item \textbf{Temporal, Functional and Scene Understanding.} To enable reasoning in dynamic environments, this category incorporates complex scene understanding tasks that require integrating video, images, and language (e.g., \textit{COSMOS}). It utilizes 3D embodied video data to evaluate the model's grasp of temporal sequences, object affordances, and overall scene semantics (e.g., \textit{RefSpatial}).
    
    \item \textbf{Decision Making and Task Planning.} This category targets high-level cognitive abilities for robotics. The data is designed to support decision-making in dynamic contexts (e.g., \textit{COSMOS}) and to translate language instructions into concrete actions (e.g., \textit{Where2Place}). This includes specific downstream tasks such as determining vacant spaces for placement (e.g.,  \textit{Vacant QA}) and executing object placement plans (e.g., \textit{RefSpatial}).
\end{itemize}

\subsection{Metaloop Data Selection}

\paragraph{Preserving General Capabilities with General Data.}
\label{sec:nat_vide}
To retain general-purpose reasoning while enhancing spatial competency, we leverage natural-world video data as our general data pool. We select the SpatialVID dataset \cite{wang2025spatialvid} but intentionally remove all expert-crafted labels to avoid overfitting to human-designed spatial priors. Instead, Qwen3VL-Plus \cite{qwen3-vl} generates 24 spatially grounded QA pairs for each of the 75k videos, covering versatile skills such as counting, relative comparison, coarse geometry, and global layout reasoning.

We then filter this data using InternVL3.5-38B \cite{internvl3.5}, which infers answers for each question twice. A QA pair is kept if InternVL3.5's answer matches the Qwen3VL-Plus answer. If InternVL3.5's two inferences are identical to each other but differ from the original, we replace the original answer with the new consensus answer. This refinement process yields 14k QAs. We further enhance this dataset with the 19k QA video subset from InternSpatial \cite{deng2025internspatial}, which also consists of pure video and text instructions without box or mask annotations.

\paragraph{Enhancing Embodied Abilities from Weakness Data.} Each loop of Metaloop requires injecting distinct weakness data that is more carefully curated and contextually aligned with the model’s evolving capabilities tailored to the model’s current capability bottlenecks. This targeted data injection ensures that each iterative training phase addresses specific underperforming areas rather than general retraining. Taking the first Metaloop round as an example, after RL training we perform four rounds of rollout inference on the SFT data pool using the post-RL model. Rule-based filtering is applied to preliminarily identify weakness samples, followed by format unification. To remove low-quality or semantically ambiguous cases, the data are scored by Qwen3VL-Plus \cite{qwen3-vl} and InternVL3.5-38B \cite{internvl3.5}, and high-quality weakness samples are selected via a voting strategy. To further ensure quality, we additionally conduct random human review to guarantee the reliability of the selected weakness samples while preserving rigorous evaluation standards.

\section{An Explanation of DPPO: Unified Preference Learning}\label{sec:expl_sub}
The theoretical underpinning of our framework is the unification of SFT and RL into a single, cohesive paradigm of Preference Learning (PL). We posit that these seemingly disparate training methodologies can be viewed as specific instantiations of a single, universal objective: maximizing the log-likelihood of observed preference data under a policy-conditioned probabilistic model.

\paragraph{The Unified Objective Function.} Let $\pi_\theta$ be a policy parameterized by $\theta$. We define a general preference dataset $D_{pref}$ composed of preference samples $\{c_i\}$. Each sample $c$ represents any form of preference expression, such as an expert trajectory, a win/loss pair, or a ranked list of outcomes. The universal objective is to find the optimal policy parameters $\theta^*$ that maximize the expected log-likelihood of this preference data:
\begin{equation}
\theta^* = \arg \max_{\theta} \mathbb{E}_{c\sim D_{pref}}[\log P(c|\pi_\theta)]
\end{equation}
The core difference between various PL algorithms lies in (1) the specific structure of the preference sample $c$, and (2) the choice of the probabilistic model $P(c|\pi_\theta)$ that links the observed preference to the agent's policy. We follow recent work in assuming that preferences are governed by a latent reward function $r(\tau)$, which can be implicitly defined by the policy itself relative to a reference policy $\pi_{ref}$, such that $r(\tau) = \beta \log(\pi_\theta(\tau)/\pi_{ref}(\tau))$.

We now demonstrate that SFT and GRPO are elegant special cases of this unified objective.

\paragraph{SFT.} Preference Sample $c$: The sample $c$ is a single expert trajectory $\tau^*$, which is considered axiomatically optimal. Thus, $D_{pref} = D_{SFT} = \{\tau_i^*\}$. Probabilistic Model $P(c|\pi_\theta)$: The implicit assumption in SFT is that the probability of observing the expert's preference for $\tau^*$ is directly proportional to the policy's likelihood of generating that trajectory.Substituting this into the unified objective reveals that maximizing the log-likelihood of the preference data is equivalent to minimizing the standard negative log-likelihood SFT loss $\mathcal{L}_{SFT}(\theta)$:

\begin{equation}
\log P(\tau^* | \pi_{\theta}) = \log P(\tau^*; \theta) = \sum_{(s, a^*) \in \tau^*} \log \pi_{\theta}(a^*|s)
\end{equation}

\paragraph{Group Relative Policy Optimization.} Preference Sample $c$: The sample $c$ is a ranked list of trajectories $\{\tau_1, \tau_2, \dots, \tau_k\}$, where $\tau_i  > \tau_{i+1}$. Probabilistic Model $P(c|\pi_\theta)$: A Plackett-Luce model is employed to define the probability of observing this specific ranking. This is the product of sequential selection probabilities:
\begin{equation}
\begin{aligned}
P(c|\pi_\theta) &= \prod_{i=1}^k P(\tau_i \text{ chosen from } \{\tau_i, \dots, \tau_k\}|\pi_\theta) \\
&= \prod_{i=1}^k \frac{\exp(r(\tau_i))}{\sum_{j=i}^k \exp(r(\tau_j))}
\end{aligned}
\end{equation}

Substituting the implicit reward function $r(\tau)$ and maximizing the log-probability of this expression recovers the GRPO objective. This demonstrates that GRPO naturally extends preference optimization to richer, rank-based preference signals.

\paragraph{Synergistic Roles of SFT and RL from the Unified Formulation.} This unified mathematical framework provides a principled explanation for the synergy between SFT and RL (with RL instantiated here as GRPO). 
(1) SFT for Knowledge Enhancement. As shown in its formulation, the SFT objective is optimized on the dataset $D_{SFT}$ containing only positive exemplars ($\tau^*$). Consequently, the gradient of the unified objective function is exclusively directed towards increasing the likelihood of these expert behaviors. This makes SFT a direct and stable mechanism for knowledge enhancement, rapidly grounding the policy in a region of demonstrated competence and instilling a foundational understanding of the task.
(2) RL for Weakness Detection and Refinement. In contrast, RL-based methods like GRPO operate on preference samples $c$  that include both superior and inferior outcomes. The unified objective promotes preferred behaviors while suppressing dispreferred ones, effectively driving weakness detection and refinement. This allows the policy to learn from its own suboptimal variations and correct subtle flaws that SFT alone may overlook.
\section{Experiment Details}\label{sec:exp_suppl}
\subsection{Benchmark Analysis}\label{sec:bench_suppl}
Current embodied benchmarks are too coarse-grained, often providing only task-level Pass/Fail metrics. This ``black-box'' evaluation is insufficient for fine-grained analysis and, critically, cannot guide iterative model updates, such as the targeted enhancement of a specific capability dimension.

To address this, we re-annotate multiple public embodied datasets according to a new taxonomy of core capability dimensions (e.g., spatial reasoning, physical causality, as detailed in Tab.~\ref{tab:benchmark}). This fine-grained taxonomy enables us to conduct a novel, diagnostic analysis. We use it to analyze the capability profile of various existing models, revealing systemic gaps. We also analyze our own Pelican-VL's capability dimensions before and after our DPPO training, allowing us to precisely measure and demonstrate the targeted impact of our self-refine framework. 

\paragraph{Capability Taxonomy.}To systematically characterize embodied understanding, we define a unified taxonomy consisting of nine primary dimensions that span the full perceptual–cognitive–action hierarchy. 
(1) Physical \& Causal Reasoning involves predicting physical feasibility, stability, and interactions such as collisions or support relations. 
(2) Perception \& Object Grounding measures visual grounding abilities including point selection, bounding box localization, and grasp-point identification. 
(3) Quantitative \& Numerical Reasoning assesses a model’s understanding of discrete counts, continuous quantities (e.g., size, length, distance), and relative magnitudes (e.g., heavier/lighter). 
(4) Spatial \& Geometric Reasoning captures spatial layout comprehension across reference frames and topological relations such as inside/outside or adjacency. 
(5) Temporal \& Sequential Reasoning evaluates the ability to infer temporal order, state transitions, and motion dynamics across frames or events. 
(6) Affordance \& Function Reasoning focuses on understanding how objects can be manipulated, opened, poured, or used as tools under physical and kinematic constraints. 
(7) Multi-Object \& Scene Consistency examines relational coherence among multiple entities and ensures spatial–semantic consistency within complex scenes. 
(8) Scene \& Action Understanding targets high-level perception, including scene categorization, human or agent action recognition, state detection, and pose or gesture interpretation. 
(9) Decision \& Task Planning addresses goal-directed reasoning, including next-action prediction, conditional if–then logic, sequential plan generation, and sub-goal decomposition. 
Together, these nine dimensions encompass the fundamental reasoning spectrum required for real-world embodied intelligence.

\paragraph{Re-labeling Analysis.} \label{sec:relab}
We re-annotated 27,667 samples from ten public embodied datasets according to a unified taxonomy comprising nine primary reasoning dimensions. The overall distribution reveals a substantial imbalance between the embodied abilities. Spatial and Geometric Reasoning dominates with 8,465 samples (30.6\%), followed by Quantitative and Numerical Reasoning with 6,142 (22.2\%) and Scene and Action Understanding with 5,359 (19.4\%). In contrast, Physical and Causal Reasoning (979, 3.5\%), Affordance and Function Reasoning (632, 2. 3\%) and Decision and Task Planning (914, 3.3\%) are severely underrepresented, while Perception and Object Grounding (1,047, 3.8\%), Temporal and Sequential Reasoning (2,910, 10. 5\%), and Multiobject and Scene Consistency (1,219, 4.4\%) occupy intermediate proportions. This uneven distribution indicates that existing embodied benchmarks disproportionately emphasize spatial, numerical, and semantic reasoning, while key competencies essential for real-world robotic intelligence, such as physical feasibility, affordance perception, and long-horizon decision planning, are sparsely represented. These findings highlight the need to develop a balanced benchmark that equitably evaluates the perceptual, physical, and cognitive dimensions of embodied intelligence.

\paragraph{Limitations of Existing Embodied Benchmarks.}\label{sec:comp}
As shown in Figure 7, the comparison across ten mainstream embodied benchmarks reveals a pronounced disparity in capability coverage. Most existing datasets concentrate on a subset of embodied reasoning—particularly Spatial \& Geometric, Quantitative, and Scene \& Action Understanding—while offering limited representation of physically grounded and goal-oriented reasoning dimensions. Physical \& Causal and Affordance \& Function reasoning appear only sporadically across benchmarks, and Decision \& Task Planning is scarcely covered beyond a few simplified next-action tasks. Temporal and sequential reasoning is also weakly represented despite its importance for dynamic perception and long-horizon control. These gaps collectively indicate that current benchmarks insufficiently capture the full perceptual–cognitive–motor hierarchy required for embodied intelligence. In contrast, our benchmark achieves comprehensive coverage across all nine primary reasoning dimensions, enabling more systematic evaluation of perception, reasoning, and decision-making competencies in embodied agents. Having established the comprehensive coverage of our proposed benchmark, we next employ it to evaluate model performance and validate its effectiveness in revealing embodied reasoning disparities across different vision-language models.

\begin{figure}
    \begin{subfigure}[b]{0.5\textwidth}
        \centering
        \includegraphics[width=\linewidth]{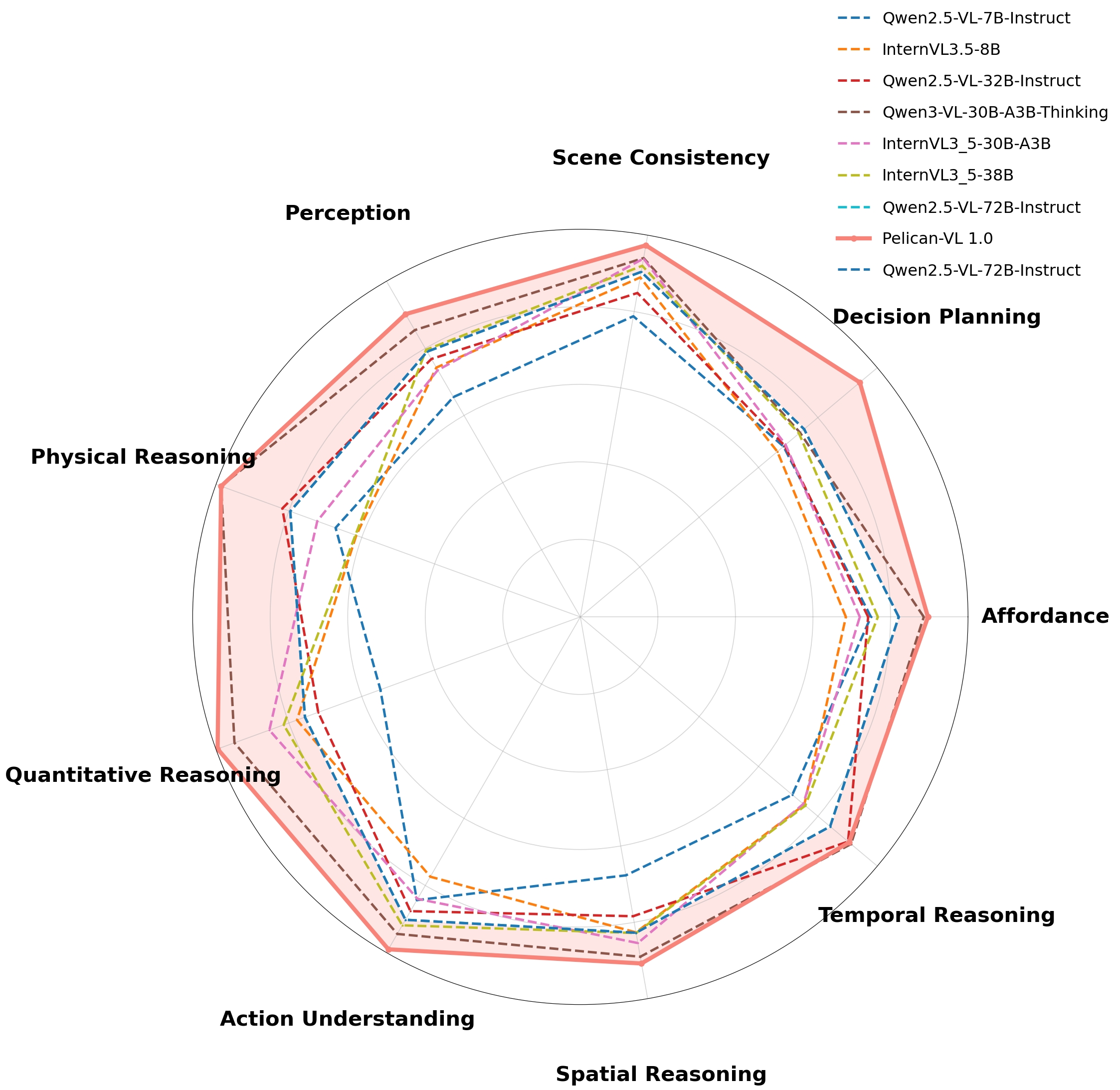}
        \caption{Pelican-VL 1.0 vs models $\leq$100B}
        \label{fig:radar_leq72b}
    \end{subfigure}
    \begin{subfigure}[b]{0.5\textwidth}
        \centering
        \includegraphics[width=\linewidth]{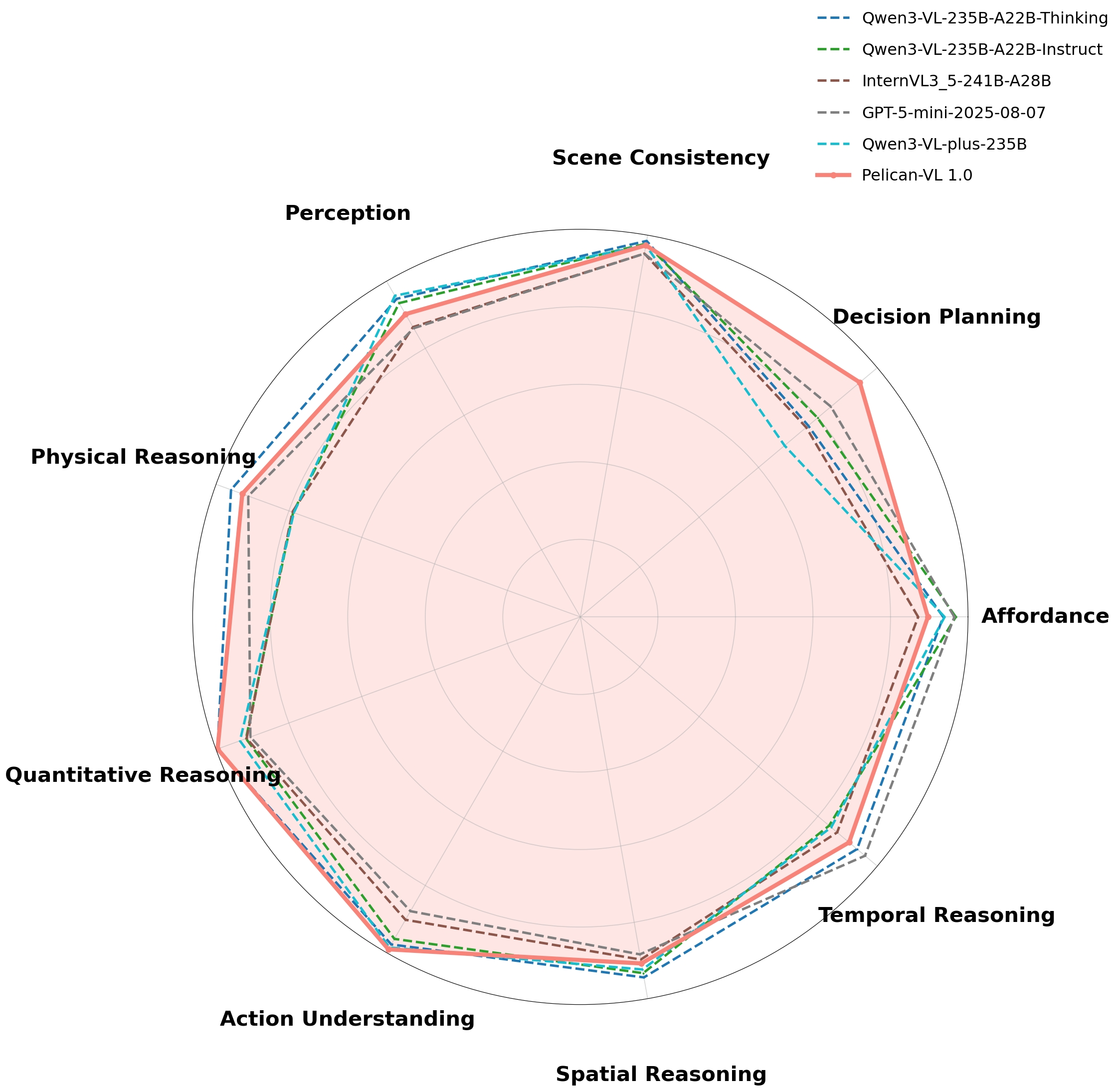}
        \caption{Pelican-VL 1.0 vs models $>$100B}
        \label{fig:radar_gt72b}
    \end{subfigure}

    \caption{Benchmark performance radar comparison of Pelican-VL 1.0 (72B) against other models across nine dimensions.}
    \label{fig:radar_comparison}
\end{figure}

\paragraph{Performance Comparison.} \label{sec:performance}
Our analysis using our new-built 9-dimension capaliblity taxonomy (Fig.~\ref{fig:radar_comparison}) reveals three critical insights. First, we observe that existing open-source models exhibit uneven capability profiles. For example, models like Qwen2.5-VL-72B-Instruct show deficits in core embodied areas like Physical Reasoning and Quantitative Reasoning, even while performing well in others like Scene Consistency. This highlights that standard VLM training fails to build holistic embodied intelligence. Second, our DPPO framework is effectively solves this. Starting from a base model with similar imbalances, Pelican-VL 1.0 achieves both a comprehensively balanced capability profile and achieves SOTA performance across all nine dimensions. Third, while large-scale, proprietary models (e.g., Qwen3-VL-plus-235B) are generally more well-rounded, our Pelican-VL 1.0 not only matches their balance but also outperforms all counterparts in critical dimensions like Decision and Task Planning and Scene and Action Understanding.

Building upon this unified evaluation framework, we systematically compare Pelican-VL 1.0 with a wide range of vision-language models. As illustrated in Fig.~7, Pelican-VL 1.0 demonstrates substantial and well-distributed performance gains over the Qwen2.5-VL-72B-Instruct base model, achieving an average improvement of 15.4\% across all nine reasoning dimensions. The most pronounced enhancements emerge in Quantitative and Numerical Reasoning, Physical and Causal Reasoning, and Decision and Task Planning, indicating that embodied, robot-centric fine-tuning effectively strengthens physical grounding, causal reasoning, and long-horizon planning. Steady improvements are also observed in Perception and Object Grounding, Scene Consistency, and Spatial Reasoning, suggesting enhanced perceptual-cognitive integration through multimodal alignment. When compared with models of similar or smaller scale (\(\leq 100\)B parameters)—including the Qwen2.5-VL-7B/32B/72B, Qwen3-VL-30B, and InternVL 3.5 series—Pelican-VL 1.0 maintains superior and balanced performance across all embodied dimensions. Against large-scale and closed-source systems exceeding 200B parameters (e.g., Qwen3-VL-235B, InternVL 3.5-241B, GPT-5-mini, and Qwen3-VL-plus-235B), our model remains comparable overall, outperforming all counterparts in Decision and Task Planning and Scene and Action Understanding. 

Overall, the consistent improvements across all reasoning dimensions validate both the robustness of Pelican-VL 1.0 and the effectiveness of our proposed benchmark. These findings demonstrate that an embodiment-aligned evaluation framework provides a reliable and holistic measure of embodied reasoning, bridging the continuum from low-level perception and physical understanding to high-level planning and decision-making. This unified perspective establishes a strong foundation for advancing embodied intelligence research and benchmarking future multimodal foundation models.

\paragraph{Comparison of Random} \label{sec:random_comparison}
Table~\ref{tab:7b_training_random_cycle} reports the results of training models
with the same data budget as DPPO, but using \emph{randomly sampled} data rather
than difficulty-aware sampling. This random strategy does not balance sample
difficulty, making the model highly susceptible to \textit{RL hacking} during
the RL stage. As a result, the model collapses into producing only final answers
and loses its chain-of-thought (CoT) reasoning ability---denoted by ``--'' in
the table. Overall, the table demonstrates that removing difficulty-aware selection
destabilises the RL–SFT process and leads to significant performance degradation
across benchmarks.

\begin{table}[htbp]
\centering
\caption{
Performance of the 7B model trained with randomly sampled data in cycle. “–” indicates that during the second RL stage, the model could only output final answers and lost its chain-of-thought (CoT) capability.}
\label{tab:7b_training_random_cycle} 
\small 
\setlength{\tabcolsep}{3pt} 
\begin{tabular}{lccccccc}
\toprule
\textbf{Cycle} & \textbf{Model} & \textbf{W2P} & \textbf{VSI} & \textbf{MVBench} & \textbf{Omni} & \textbf{Emb} \\
\midrule
& Base & 24.0 & 37.2 & 68.1 & 39.2 & 70.4 \\
\midrule
\multirow{2}{*}{Stage 1}   & RL   &25.1 & 44.1 & 68.4 & 43.8 & 70.8 \\
 & SFT  &35.4& 50.1 & 68.2 & 44.9 & 69.9\\
\midrule
Stage 2 & RL   & - & - & - & - & -  \\
\bottomrule
\end{tabular}
\end{table}

\paragraph{Comparison of Chain-of-Thought Reasoning Ability.} \label{sec:cot_comparison}
To assess the models’ reasoning behavior, Table~\ref{tab:7b_training_cot} and 
Table~\ref{tab:7b_training_log_single_col} present performance under two 
evaluation modes: \emph{step-by-step chain-of-thought (CoT) reasoning} and 
\emph{direct answer prediction}. Table~\ref{tab:7b_training_cot} tracks the 
evolution of CoT capability across training stages, while 
Table~\ref{tab:7b_training_log_single_col} reports the corresponding performance 
when the models produce only final answers.Notably, by stage~3, the direct-answer performance shows only marginal 
improvement, yet the CoT accuracy continues to increase by more than 5\% on 
several datasets. This consistent gain in reasoning ability indicates that the 
model’s generalization is still being strengthened, highlighting the necessity 
and effectiveness of running all three DPPO Metaloop cycles.

\begin{table}[htbp]
\centering
\caption{Comparison results between Pelican-VL 7B and other models under the chain-of-thought (CoT) reasoning evaluation.}
\label{tab:7b_training_cot}
\small 
\setlength{\tabcolsep}{3pt} 
\begin{tabular}{llccccccc}
\toprule
\textbf{Cycle} & \textbf{Model} & \textbf{W2P} & \textbf{VSI} & \textbf{Cosmos} & \textbf{Ego.}  \\
\midrule
& Base & 11.6 & 30.3  & 51.0 & 60.6 \\
& Only RL & 36 & 44.9  & 51.5 & 67.4  \\
& Only SFT & 25.4 & 40.7  & 51.0 & 59.2  \\
\midrule
\multirow{2}{*}{Stage 1}   & RL   &44.5 & 35.0  & 40.3 & 61.4  \\
 & SFT  &35.4& 40.2  & 49.5 & 70  \\
\midrule
\multirow{2}{*}{Stage 2} & RL   &41.2 & 42.6  & 52.0 & 70.6   \\
 & SFT  &54.0 & 45.3  & 52.6 & 65.6  \\
\midrule
Stage 3 & RL   &59.0 & 52.4  & 51.1  & 68.8  \\
\bottomrule
\end{tabular}
\end{table}

\begin{table}[t]
\centering
\caption{Performance evolution across the stages of Pelican-VL 7B. The model exhibits consistent improvements throughout all phases, with Phase 3 yielding additional enhancements in COT reasoning and generalization ability. }
\label{tab:7b_training_log_single_col}
\small 
\setlength{\tabcolsep}{3pt} 
\begin{tabular}{llccccc}
\toprule
\textbf{Cycle} & \textbf{Model} & \textbf{W2P} & \textbf{VSI} & \textbf{MVBench} & \textbf{Omni} & \textbf{Emb} \\
\midrule
& Base & 24.0 & 37.2 & 68.1 & 39.2 & 70.4 \\
\midrule
\multirow{2}{*}{Stage 1}   & RL   &32.2 & 39.9 & 68.0 & 40.6 & 72.4 \\
 & SFT  &38.6& 47.7 & 68.0 & 43.8 & 72.9\\
\midrule
\multirow{2}{*}{Stage 2} & RL   &43.4 & 48.3 & 68.3 & 45.2 & 73.2  \\
 & SFT  &49.5 & 53.8 & 68.1 & 46.3 & 74.6  \\
\midrule
Stage 3 & RL   &53.5 & 54.3 & 67.8&  46.5 & 75.5 \\
\bottomrule
\end{tabular}
\end{table}

\begin{table*}[h]
\footnotesize
\setlength{\tabcolsep}{3pt}
\renewcommand{\arraystretch}{1.25}
\centering
\caption{Task examples for each dimension in our benchmark.}
\begin{tabular}{>{\raggedright}p{2.8cm} >{\raggedright}p{3.7cm} >{\raggedright\arraybackslash}p{7.6cm}}
\toprule
\textbf{Primary dimensions} & \textbf{Secondary dimensions} & \textbf{Example} \\
\midrule
\textbf{Physical \& Causal Reasoning} & Physical feasibility  & Can the tissue box fit behind the chair? — Yes. \\
 & Stability and support  & Will the stacked cups remain stable if the top one is pushed slightly? — No. \\
 & Intuitive physics and collisions & If the rolling ball hits the block, will it fall over? — Yes. \\
\midrule
\textbf{Perception \& Object Grounding} & Click or coordinate prediction & Please point out the orange box. — (x, y) = (0.43, 0.72). \\
 & Bounding box detection & Identify the bounding box of the cup on the table. — (x1, y1, x2, y2). \\
 & Grasp point & Where should the robot grasp the handle of the kettle? — At the handle center. \\
\midrule
\textbf{Quantitative \& Numerical Reasoning} & Counting & How many apples are on the table? — Three. \\
 & Continuous estimation & What is the distance between the box and the wall? — About 0.5 meters. \\
 & Relative magnitude & Which container holds more water? — The right one. \\
\midrule
\textbf{Spatial \& Geometric Reasoning} & Spatial relations & Can the speaker fit left of the cup? — Yes. \\
 & Egocentric vs allocentric frames & From the robot’s view, which object is on its right? — The red box. \\
 & Topological relations & Which object is inside the basket? — The towel. \\
\midrule
\textbf{Temporal \& Sequential Reasoning} & Temporal order & What happened before the man picked up the ball? — He bent down. \\
 & State transition and action effects & After the cup is tilted, what happens to the water? — It spills. \\
 & Motion and sequence reasoning & What is the person doing after opening the door? — Walking inside. \\
\midrule
\textbf{Affordance \& Function Reasoning} & Graspability & Which part of the screwdriver should be grasped? — The handle. \\
 & Openability or pourability & Can the bottle be opened by twisting the cap? — Yes. \\
 & Tool-use reasoning & What tool can be used to cut the rope? — Scissors. \\
 & Manipulation feasibility & Is it possible to rotate the knob with one hand? — Yes. \\
\midrule
\textbf{Multi-Object \& Scene Consistency} & Multi-object relations & Which object is partially hidden behind the monitor? — The keyboard. \\
 & Scene-state consistency & Is the cup in the same place across both frames? — Yes. \\
 & Spatial–language consistency & Does the description match the image scene? — No. \\
\midrule
\textbf{Scene \& Action Understanding} & Scene types & What type of environment is this image captured in? — Kitchen. \\
 & Action recognition & What is the person doing in the video? — Pouring water. \\
 & State recognition & Is the fridge door open or closed? — Open. \\
 & Pose, gesture, direction & What direction is the person pointing to? — Left. \\
\midrule
\textbf{Decision \& Task Planning} & Next-action prediction & What should the robot do after picking up the cup? — Place it on the tray. \\
 & Sequential plan & List the steps to make a cup of tea. — Boil water → Add tea → Pour → Serve. \\
 & Sub-goal decomposition & What are table assembly sub-goals? — Align legs → Screw bolts → Tighten. \\
\bottomrule
\end{tabular}
\label{tab:benchmark}
\end{table*}

\begin{table*}[h]
\small
\caption{Coverage of the nine primary embodied reasoning dimensions across mainstream benchmarks and our proposed dataset.}
\centering
\resizebox{\textwidth}{!}{
\begin{tabular}{>{\raggedright\arraybackslash}p{4.2cm} | *{8}{c}}
\toprule
\textbf{Primary dimensions} & \textbf{COSMOS}& \textbf{ERQA}& \textbf{Phyx}& \textbf{W2P}& \textbf{VSI-Bench}& \textbf{BLINK}& \textbf{RoboSpatial}& \textbf{EmbSpatial} \\
\midrule
Physical \& Causal Reasoning & \checkmark & \checkmark & \checkmark & \phantom{\checkmark} & \checkmark & \checkmark & \checkmark & \checkmark \\
Perception \& Object Grounding & \phantom{\checkmark} & \checkmark & \phantom{\checkmark} & \checkmark & \phantom{\checkmark} & \checkmark & \checkmark & \phantom{\checkmark} \\
Quantitative \& Numerical Reasoning & \phantom{\checkmark} & \checkmark & \checkmark & \phantom{\checkmark} & \checkmark & \checkmark & \phantom{\checkmark} & \checkmark \\
Spatial \& Geometric Reasoning & \checkmark & \checkmark & \checkmark & \checkmark & \checkmark & \checkmark & \checkmark & \checkmark \\
Temporal \& Sequential Reasoning & \checkmark & \checkmark & \checkmark & \phantom{\checkmark} & \checkmark & \checkmark & \phantom{\checkmark} & \phantom{\checkmark} \\
Affordance \& Function Reasoning & \checkmark & \checkmark & \checkmark & \checkmark & \checkmark & \checkmark & \phantom{\checkmark} & \phantom{\checkmark} \\
Multi-Object \& Scene Consistency & \checkmark & \checkmark & \checkmark & \phantom{\checkmark} & \checkmark & \checkmark & \phantom{\checkmark} & \checkmark \\
Scene \& Action Understanding & \checkmark & \checkmark & \checkmark & \phantom{\checkmark} & \checkmark & \checkmark & \checkmark & \checkmark \\
Decision \& Task Planning & \checkmark & \checkmark & \phantom{\checkmark} & \phantom{\checkmark} & \checkmark & \phantom{\checkmark} & \phantom{\checkmark} & \phantom{\checkmark} \\
\bottomrule
\end{tabular}
}
\label{tab:coverage}
\end{table*}

\end{document}